\documentstyle[fullpage,authdate,fleqn]{report}

\def\thebibliography#1{\chapter*{Bibliography}
  \addcontentsline
  {toc}{chapter}{Bibliography}\list
  {\relax}{\setlength{\labelsep}{0em}%
\setlength{\itemindent}{-\bibhang}%
\setlength{\leftmargin}{\bibhang}}
    \def\newblock{\hskip .11em plus .33em minus -.07em}
    \sloppy
    \sfcode`\.=1000\relax}

\newcommand{\eg}{e.g., }
\newcommand{\ie}{i.e., }
\newcommand{\be}{\begin{equation}}
\newcommand{\ee}{\end{equation}}
\newcommand{\beq}{\begin{eqnarray}}
\newcommand{\eeq}{\end{eqnarray}}
\newcommand{\nl}{\nonumber \\}
\newcommand{\dc}{$^\circ$C}

\begin{document}

\pagenumbering{roman}
\pagestyle{empty}
\null
\vfill

\begin{centering}
{\huge Deriving Procedural and Warning \\
\vspace*{1mm}
Instructions from Device and \\
\vspace*{4.4mm}
Environment Models} \\
\vspace*{25mm}
by \\
\vspace*{25mm}
{\LARGE Daniel Ansari} \\
\vspace*{40mm}
Department of Computer Science \\
University of Toronto \\
Toronto, Canada \\
June 1995 \\
\vspace*{10mm}
A thesis submitted in conformity with the requirements \\
for the degree of Master of Science at the \\
University of Toronto \\
\vspace*{15mm}
Copyright \copyright\ 1995 by Daniel Ansari \\
\end{centering}

\vfill

\newpage
\pagestyle{plain}
\vspace*{2cm}
\begin{center}\Large\bf Abstract \end{center}

\noindent There has been much interest lately in the automatic generation of
documentation; however, much of this research has not considered the
cost involved in the production of the natural language generation
systems to be a major issue: the benefits obtained from automating the
construction of the documentation should outweigh the cost of
designing and coding the knowledge base.

This study is centred on the generation of {\it instructional text},
as is found in instruction manuals for household appliances.  We show
how knowledge about a device {\it that already exists as part of the
engineering effort\/}, together with adequate, domain-independent
knowledge about the environment, can be used for reasoning about
natural language instructions.

The knowledge selected for communication can be planned for, and all
the knowledge necessary for the planning should be contained (possibly
in a more abstract form) in the knowledge of the artifact together
with the world knowledge.  We present the planning knowledge for two
example domains, in the form of axioms in the {\it situation
calculus}.  This planning knowledge formally characterizes the
behaviour of the artifact, and it is used to produce a basic plan of
actions that both the device and user take to accomplish a given goal.
We explain how the instructions are generated from the basic plan.
This plan is then used to derive further plans for states to be
avoided.  We will also explain how warning instructions about
potentially dangerous situations are generated from these plans.
These ideas have been implemented using Prolog and the Penman natural
language generation system.

Finally, this thesis makes the claim that the planning knowledge
should be {\it derivable\/} from the device and world knowledge; thus
the need for cost effectiveness would be met.  To this end, we suggest
a framework for an integrated approach to device design and
instruction generation.

\newpage
\vspace*{2cm}
\begin{center}\Large\bf Acknowledgments \end{center}

\noindent First of all, I wish to thank my supervisor, Dr.\ Graeme Hirst,
for his valuable criticisms, for his advice when it was much needed, for
all the time spent reading my work, and for enduring my productivity mood
swings.

Many thanks go to Phil Edmonds, Yves Lesp\'{e}rance, and my second reader
Dr.\ Jeffrey Mark Siskind, for their helpful comments on an earlier draft
of this thesis.

I gratefully acknowledge the financial support provided by Science and
Engineering Research Council (U.K.) grant 92600436, and use of the
equipment provided by Natural Science and Engineering Research Council of
Canada.

I also wish to thank my family who, although thousands of miles away across
the ocean, gave me lots of support.

Finally, my deepest gratitude goes to my darling Sonia, who provided me
with much love, companionship, and encouragement.

\newpage
\tableofcontents

\newpage
\listoftables

\newpage
\pagenumbering{arabic}
\chapter{Introduction}
\label{ch:introduction}

Until recently, natural language generation (NLG) has been of interest mostly
to academic researchers, but applications based on this technology have
started to emerge in industry (\eg Advanced Technologies Applications,
Inc.\ \shortcite{docexpress}, Goldberg et al.\ \shortcite{goldbergetal94}).
There has been much interest lately in the automatic generation of
documentation, in particular, system and software engineering documentation
(\eg Advanced Technologies Applications, Inc.\ \shortcite{docexpress}),
technical documentation (\eg Reiter et al.\ \shortcite{reiteretal95},
R\"{o}sner and Stede \shortcite{rosner&stede94}), and instructional text (\eg
Feiner and \mbox{McKeown} \shortcite{feiner&mckeown90}, Wahlster et
al.\ \shortcite{wahlsteretal93}).  However, much of the research has not
considered the cost involved in the production of the NLG systems to be a
major issue.  This consideration is much the same as that of trying to
minimize the cost of producing an interlingua for a multi-lingual NLG
system\footnote
     {TECHDOC \cite{rosner&stede94} is one example of a multi-lingual
     technical documentation generation system.}:
the benefits obtained from automating the construction of the documentation
should outweigh the cost of designing and coding the interlingua, or knowledge
base.

The IDAS project of Reiter et al.\ \shortcite{reiteretal92,reiteretal95}
serves as a key motivation for our work.  One of the primary goals of the IDAS
project was to automatically generate technical documentation from a domain
knowledge base containing design information (such as that produced by an
advanced computer-aided design tool) using NLG techniques.  IDAS turned out to
be successful in demonstrating the usefulness, from a cost and benefits
perspective, of applying NLG technology to partially automate the generation
of documention.  This technical documentation was intended to be read by
technicians and other experts, so the focus of this work is different from
ours.

This study is centred on the generation of {\it instructional text}, as is
found in instruction manuals for household appliances.  We will endeavour to
show how knowledge about a device {\it that already exists as part of the
engineering effort\/}, together with adequate, domain-independent knowledge
about the environment, can be used for generating natural language
instructions.  We will also describe how all this knowledge can be used for
generating warning instructions, \ie cautions to the user directing them to
avoid certain situations.

As we will see in section~\ref{sec:paris&scott}, part of an instruction manual
may contain {\it safety information}, or this information might accompany text
given in the {\it use\/} part of the manual if it is specific to a particular
step in achieving a task.  It is our view that the knowledge relevant to
warnings and safety advice is naturally not closely tied to the sequence in
which the steps should be carried out, but is more concerned with the {\it
consequences\/} of not carrying out the steps in an appropriate manner, and
more generally with consequences of handling the appliance incorrectly.
Hence, it is possible for certain types of knowledge to be used for generating
text about safety and warnings.

Delin et al.\ \shortcite{delinetal93} suggested that it is useful to
distinguish six levels of representation of instructional texts:
\begin{enumerate}
     \item {\bf The knowledge of the artifact}$\:$ A functional model of the
     artifact and its mode of operation in terms of actions and states.
     \item {\bf The deep intentions}$\:$ The representation of the
     originator's intention that the user perform the sequence of actions that
     constitute a particular task involving the artifact.
     \item {\bf The knowledge selected for communication}$\:$ What is to be
     communicated about the artifact and the task that will enable the user to
     perform the appropriate actions, making assumptions about their cultural
     background, world knowledge, and expertise.
     \item {\bf The shallow intentions}$\:$ A representation of the goals that
     the text has to achieve in order to motivate the required tasks.
     \item {\bf The rhetorical structure}$\:$ The discourse strategies chosen
     to achieve the text's goals.
     \item {\bf The syntactic structure}$\:$ The syntax expressing the chosen
     rhetorical structure.
\end{enumerate}

As we shall see in chapter 2, the systems that researchers have built so far
to generate instructional text have largely ignored representation level~1,
and most have assumed the prior existence of level~3 knowledge.

We claim that the {\it deep intentions\/} can be encoded in the {\it world
knowledge}, which should include knowledge about the environment, in
particular the way a human agent interacts with general types of components
such as buttons, levers, and lids.  For example, the fact that a button must
be pressed in order for a circuit to be completed is part of the knowledge
about the artifact.  The fact that in order for the button to become pressed
the user can perform the {\it press\/} action on this button should be
represented in the world knowledge: it is a general fact that applies to any
button, and it is the intention of the originator of the instructions that the
button be pressed by the user.

The knowledge selected for communication can be planned for, and all the
knowledge necessary for the planning should be contained (possibly in a more
abstract form) in the knowledge of the artifact together with the world
knowledge.  The kinds of device and world knowledge that should be sufficient
for this planning will be discussed in chapter~3.

In chapter~4 we shall present the planning knowledge for two example
domains---a toaster and a breadmaker---in the form of axioms in the {\it
situation calculus}.  This planning knowledge formally characterizes the
behaviour of the artifact, and it is used to produce a basic plan of actions
that both the device and user take to accomplish a given goal.  The axioms
together with the goal are the input to our system.  We will explain how the
instructions are generated from the basic plan.  This plan is then used to
derive further plans for states to be avoided.  We will also explain how
warning instructions about potentially dangerous situations are generated from
these plans.  Thus, the output of our system consists of English natural
language instructions, including warning instructions, for how to use the
device to achieve its purpose.

We make the assumption that the device and world knowledge take the form of
formal specifications.  This thesis also makes the claim that the planning
knowledge should be {\it derivable\/} from the device and world knowledge;
thus the need for cost effectiveness would be met.  We shall attempt to
justify this claim, to some extent, in chapter~5.  However, this is such a
difficult problem that we do not expect a solution to be found in the near
future.

Finally, we will suggest a framework for an integrated approach to device
design and instruction generation.  We will also discuss directions for future
work.

The main contributions of this thesis are the following:
\begin{enumerate}
     \item the suggestion that an integrated model of the device (including
     solid, kinematic, electrical, and thermodynamic models) together with
     world knowledge can be used to automate the generation of instructions,
     including warning instructions;
     \item that situations in which injuries to the user can occur need to be
     planned for at every step in the planning of the {\it normal\/} operation
     of the device, and that these ``injury sub-plans'' are used to instruct
     the user to avoid these situations.  Thus, unlike other instruction
     generation systems, our system tells the reader what {\it not\/} to do as
     well as what to do; and
     \item the notion that actions are performed on the materials that the
     device operates upon, that the states of these materials may change as a
     result of these actions, and that the goal of the system should be
     defined in terms of the final states of the materials.
\end{enumerate}

\chapter{Related research}

\section{Planning for instructions}
\label{sec:agre&horswill}

Agre and Horswill \shortcite{agre&horswill94} presented an object-centred
formalization of action.  They contend that any computational theory of action
should have two properties:
\begin{enumerate}
     \item It should explain how agents can achieve goals and maintain
     background conditions\footnote{Background conditions specify that all
     instances of a given type should be in a given state and location.}.
     \item It should explain how agents can choose their actions in real time.
\end{enumerate}
They proposed that part of the solution for achieving these properties of
correctness and efficiency lie in {\it culture}, and specifically in {\it the
formal properties of a given culture's repertoire of artifacts}.  They defined
an interesting class of tasks, called {\it cooking tasks}, as tasks which only
involve objects in certain classes, and implemented a program, Toast, which
demonstrates that cooking tasks can be planned in a ``greedy'' (without
backtracking) fashion.  The efficiency issue is addressed by constraining the
types of objects and goals that are manipulated, so that the agent can always
choose an action which will move it closer to its goal without constructing a
plan.  Agre and Horswill say that the inventory of objects available to an
agent depends upon that agent's culture, and by distinguishing the forms of
improvised activity which can be performed by simple mechanisms from the more
complex and varied, an elaborate planning paradigm is not necessary.

Agre and Horswill's formalism was created with the intent of analyzing
interactions between an agent and its environment.  They presented an outline
of a formal model of objects, actions, and tasks, defining object types,
action types, and tasks in terms of states of the objects and the world, and
goals.  A world state satisfies a goal in their formalism if that state
includes some instance of the indicated type that is in the indicated state.
By categorizing object types in terms of the properties of their {\it state
graphs}, they defined a {\it cooking task} as a task which involves only {\it
tools\/} and {\it materials}.  According to Agre and Horswill, these two
object classes, together with {\it containers}\footnote
     {Examples of tools are forks, spoons, and knives; examples of materials
     are pancake batter, milk, eggs, and bread slices; examples of containers
     are bowls, cups, and plates.},
constitute the vast majority of objects found in the average kitchen.

They described an algorithm that solves a cooking task and sketched the design
of an agent which can carry out this process.  Their general idea is that the
agent is in the kitchen and can readily detect the states of all visible
objects.  The agent achieves its goals by performing actions using the
tools\footnote
     {Each tool has its own set of states and actions, since it is also an
     object.}
to push materials through certain customary state transitions\footnote
     {These are defined by a {\it state graph}.}.
Their algorithm also uses an {\it action table\/} and a {\it tool table\/} to
determine what actions and/or tools can be applied in order to move towards
the goal state.  The goals are represented as triples of the form (class,
state, container) specifying that an object of the specified class in the
specified state should be placed in the specified container.

Agre and Horswill's work is interesting in that it attempts to model
interactions between an agent and its environment, which is what a system that
generates natural language instructions, particularly warning instructions,
should do to some extent.  Their observation that materials go through certain
state transitions is also relevant to the current study.  However, they do not
go any further than proposing a formalism that provides efficient planning in
cooking tasks.  Also, they do not consider the modelling of complex devices in
the kitchen, which is important for the current study.

\section{The Penman system}
\label{sec:penman}

Penman is a flexible sentence-level text generator that was developed at the
USC Information Sciences Institute \cite{mann85,matthiessen85,penman}.  It
provides a broad coverage of English syntax, probably the most comprehensive
of any readily available text generator.  Penman is based on a
systemic-functional view of language \cite{halliday76}: its approach is
functional, that is, it uses features of the context to map communicative
goals to acceptable grammatical forms.  A by-product of this view of language
is that the system contains a well-developed implementation of the {\it
systemic network}.  Penman traverses this network, which effects the
generation of sentence structures.

Penman provides two fundamental interfaces for surface realization of the
text: the SPL (Sentence Plan Language) command interface, and the raw inquiry
interface.  The latter allows one to exercise complete control over Nigel
(Penman's grammatical component), but to specify the great number of responses
required would be a tedious operation.  SPL is an extensive and flexible
language that allows the specification of sentences in terms of the processes
they are based upon, and the entities that participate in those processes.
The SPL specification is used by Penman to provide responses (including
default responses) to the various inquiries.

In order to use Penman to generate text, a domain model and lexicon must be
specified.  The Upper Model, which is provided by Penman, and the domain
model, which is defined by the user, contain definitions of the entities that
the text should address.  Both models contain a taxonomy of entities in the
world which aids the generation of English, and the domain model is linked
with the Upper Model.  The lexicon contains the definitions of words---their
spellings, variant forms, and other features.

For example, the sentence:
\begin{quote}
     Knox sails to Pearl Harbor.
\end{quote}
is specified by:
\begin{small}
\begin{verbatim}
     ((S1 / SAIL
          :actor (KNOX / SHIP)
          :destination (PEARL-HARBOR / PORT)
          :tense PRESENT
          :speechact ASSERTION))
\end{verbatim}
\end{small}
This specification describes one particular sailing action called \verb"S1"
that has \verb"KNOX" as its {\small ACTOR}, \verb"PEARL-HARBOR" as its {\small
DESTINATION}, and that this information should be asserted in the present
tense.  On its own, \verb"KNOX" is just a symbol, so we also need to tell
Penman that this symbol represents an instance of \verb"SHIP", which is a
domain model entity.

\section{Analyzing instructional text}

\subsection{Paris and Scott}
\label{sec:paris&scott}

Paris and Scott, who have been conducting work on generating multilingual
instructions, insist that computational systems should be able to generate the
variations found in texts.  Their paper \cite{paris&scott94} is one step in
this direction.

In this study, Paris and Scott described different ways, or {\it stances}, in
which instruction manuals can convey information:
\begin{description}
     \item [Information provision] Factual knowledge is provided which
     augments the reader's knowledge of the artifact or the task.
     \item [Eulogy] The text accentuates the positive aspects of the product,
     or ``congratulates'' the user for purchasing the product.
     \item [Directive] An order is given describing how the user should
     perform a task, without a rationale being given.
     \item [Explanation] The reader is given advice on how to perform a task
     together with an explanation as to why it should be performed in the
     prescribed manner.\footnote
          {An instruction conveyed by an {\it explanation\/} stance may be
          realized as a {\it matrix clause\/} together with a {\it purpose
          clause\/} \cite{dieugenio92b}.  The matrix clause describes the
          action, and the purpose clause expresses an agent's purpose in
          performing that action.}
\end{description}

Paris and Scott noted that the particular stance employed for presenting
information at any point in a manual seems to be influenced by factors such as
safety, requirements for memorability, and the expected expertise of the
reader.  Also, the forms in which each stance can be realised seems to be
determined partly by language.  For example, one instruction for filtering
coffee may be presented as a directive in English, whereas the French version
may more appropriately be given as information provision.

They found that some manuals are divided into distinct sections, as follows,
with each section typically adopting a particular stance:
\begin{description}
     \item [General information about the product] This section generally
     consists of text which congratulates the user for purchasing the product,
     describes the product and its advantages, and gives conditions of the
     warranty.  The stances adopted for this part are usually {\it information
     provision\/} and {\it eulogy}.
     \item [Information about safety, etc.] This includes warnings, general
     safety advice, and crucial steps to be performed (either to {\it
     accomplish\/} the task or to obtain {\it better\/} results).  The stance
     can be either a {\it directive\/} or an {\it explanation}.
     \item [Preparatory steps or installation] This is information on how to
     prepare the product for use.
     \item [Use] This explains how to operate the product.
     \item [Care and maintenance] This part informs the reader how to clean
     and care for the artifact.
     \item [Trouble-shooting] This part is intended to help the reader
     identify the source of any potential problems, and to provide information
     about the possible consequences of not carrying out a step properly.
     Actions to be performed to remedy the problem are provided, together with
     conditions under which they are appropriate.  The stance is usually {\it
     directive\/} (actions allowing the reader to pinpoint the problems are
     given).
\end{description}

Paris and Scott observed that non-sectioned manuals may present the above
information in an interleaved fashion, especially if space is a problem and
the writers do not wish to divide the manual into such sections.

\subsection{Vander Linden}
\label{sec:vanderlinden}

Vander Linden \shortcite{vanderlinden93b} addressed the problem of determining
the precise rhetorical and grammatical forms that are most effective for
expressing actions in an instructional context.  His major contribution to the
field of natural language processing is the application of the scientific
method for managing this diversity of expression: collecting a suitable corpus
of text, analyzing that text, implementing the results of the analysis in a
text generator, and comparing the output of the generator with the corpus.

Instructional text can be viewed as the expression of a set of actions bearing
procedural relationships with one another.  Two tasks that an instructional
text generator must perform are, first, to choose, for each action expression,
the rhetorical relation it will hold with the other actions that best conveys
their procedural relationships, and, secondly, to choose the grammatical form
that will realise this rhetorical relation.

Vander Linden did not attempt to identify the rhetorical status and the
grammatical form that appear to most effectively express various types of
actions and their relations, because it is unclear how accurate this intuitive
approach would be.  Rather, he used a detailed function-to-form study of a
corpus of instructional texts, made up of approximately 1000 clauses from 6000
words of text taken from manuals.  This corpus was represented in a relational
database representing the rhetorical and grammatical aspects of the text.

The corpus was analysed and RST (Rhetorical Structure Theory
\cite{mann&thompson86a,mann&thompson86b}) structures were built for the whole
text.  This analysis of rhetorical status made use of three nucleus-satellite
relations: {\small PURPOSE}, {\small PRECONDITION AND RESULT}, and two joint
schemas: {\small SEQUENCE} and {\small CONCURRENT}.  This set of relations and
schemas, which proved effective for the analysis, was based on the notions of
hierarchical and non-linear plans and the use of preconditions and
postconditions in automated planners.

Given this coding of the rhetorical status of action expressions, coupled with
the coding of the grammatical form of the expressions, a functional analysis
was performed which identified systematic co-variation between functions and
forms in the corpus.  It was found that a set of approximately 70 features of
the communicative environment (\ie the {\it instructional register}, in
systemic-functional terms) was sufficient to produce a broad analytical
coverage of the rhetorical status and grammatical forms used in the corpus.  A
Penman-style systemic network was used to distinguish these features and
accommodate them in a hierarchy.

Vander Linden's text generator, IMAGENE, makes decisions on the basis of
features of instructional text; it does not perform any text planning.  There
are two main inputs to IMAGENE: (1)~the structure of the process being
described (\ie the text plan), and (2)~the responses to a set of {\it
text-level inquiries}, analogous to the sentence-level inquiries of Penman.
Using these, an SPL specification is constructed, which is fed into Penman to
generate the English sentences.

The process structure is represented by a Process Representation Language
(PRL), which allows the representation of actions in a hierarchy and provides
facility for representing concurrency.  A PRL specification represents the
actions and their attributes, which have the following slots (from \cite[pages
60--61]{vanderlinden93b}):
\begin{description}
     \item [Action-Type] The lexical item corresponding to this action.
     \item [Actor] The PRL entity which represents the actor.
     \item [Actee] The PRL entity which represents the object acted upon by
     the actor.
     \item [Destination] The PRL entity which represents the destination of a
     moving action.
     \item [Duration] The natural number denoting the number of duration units
     that an action takes.
     \item [Duration-Units] The lexical item corresponding to the units of the
     duration.
     \item [Instrument] The PRL entity which represents the instrument used in
     the action.
\end{description}
In addition, the PRL entities, which represent the objects referred to by the
actions, have attributes associated with them (see \cite[page
61]{vanderlinden93b} for a listing of these).  Thus, a planner which produces
PRL structures should be able to deal with temporal information at some level,
and should be hierarchical, in order to take full advantage of IMAGENE's
expressive power.

An example of part of a PRL input is the following, in which the ``root''
action consists of an \verb"instruct" action, followed by a \verb"remove"
action, followed by a \verb"place" action:
\begin{small}
\begin{verbatim}
     (tell (:about *prl-root* Action
               (subaction instruct-action)
               (subaction remove-action)
               (subaction place-action)))

     (tell (:about instruct-action Action
               (action-type it::instruct)
               (actor phone)
               (actee hearer)))

     (tell (:about phone Object
               (object-type it::phone)
               (object-status device)))

     (tell (:about hearer Object
               (object-type it::hearer)))

     (tell (:about remove-action Action
               (subaction grasp-action)
               (subaction pull-action)
               (action-type it::remove)
               (actor hearer)
               (actee phone)))

     (tell (:about place-action Action
               (subaction return-action)
               (action-type it::place-call)
               (actor hearer)
               (actee call)))
\end{verbatim}
\end{small}
The text-level inquiries take place during the run of IMAGENE.  One example of
such an inquiry is the following, in which READER-KNOWLEDGE-Q is a question
about one particular feature of the instructional register:
\begin{small}
\begin{verbatim}
               READER-KNOWLEDGE-Q: Is INSTRUCT-ACTION a procedural sequence
               that the reader is assumed to know?
     Enter inquiry answer:
      1. KNOWN
      2. UNKNOWN
     Number Of Choice: 1
\end{verbatim}
\end{small}
The output of IMAGENE, given the full PRL and text-level inquiry inputs
corresponding to the above, is this:
\begin{quote}
     When you are instructed, remove the phone by grasping the top of the
     handset and pulling it.  Return to a seat to place a call.
\end{quote}

One type of action that IMAGENE does not handle and cannot represent in its
PRL is a {\it negative\/} action, or one that should not be performed.  If
IMAGENE is to produce warning instructions, it must be extended to deal with
these.

\section{Generating instructional text}

\subsection{Mellish and Evans}
\label{sec:mellish&evans}

Mellish and Evans \shortcite{mellish&evans89} addressed the problem of
designing a system that accepts a plan structure of the sort generated by AI
planning programs, and produces natural language text explaining how to
execute the plan.

Their system used, as input, the data structures produced by the NONLIN
hierarchical planner \cite{tate76}.  The process of natural language
generation from here can be thought of as consisting of four stages, centring
on the construction and manipulation of an expression of their {\it message
language}.  The first stage is {\it message planning}, where the generator
decides on the content and order of the real-world objects and relationships
to be expressed in language.  The output of this stage is a message language
expression.  In the {\it message simplification\/} stage, this expression is
simplified by the repeated application of localized rewrite rules.  The goal
of the next stage, {\it structure building}, is to build a functional
description of a linguistic object that will realize the intended message.
The structure-building rules are responsible for making choices from a limited
number of possible syntactic structures, introducing pronominalization where
appropriate, and accessing the lexical entries corresponding to the actions,
states, and objects.  These rules are applied as in a production system, \ie a
recursive descent traversal of the message is made.  The final stage is to
produce a linear sequence of words.

An example of an expression in the message language (before simplification)
corresponding to this piece of text follows (from \cite[page
237]{mellish&evans89}):
\begin{quote}
     If you go to the front of the car now you will not be at the cab
     afterwards.  However in order to start the engine you must be at it.
     Therefore before going to the front of the car you must start the engine.
\end{quote}
The initial message expression is:
\begin{small}
\begin{verbatim}
     implies(
        contra_seq(
           hypo_result(
              user,
              achieve(goal(located(mech, frontofcar))),
              not(goal(located(mech, cab)))),
           prereqs(
              user,
              then(wait([]), achieve(goal(started(engine)))),
              goal(located(mech, cab))))
        neccbefore(user,
           then(wait([]), achieve(goal(started(engine)))),
           achieve(goal(located(mech, frontofcar)))))
\end{verbatim}
\end{small}
where expressions such as \verb"goal(located(mech, frontofcar))" are straight
NONLIN expressions translated into Prolog.  This expression can be read
approximately as ``the hypothetical result of going to the front of the car is
that you will not be in the cab, and this contrasts with the prerequisite of
being in the cab to start the engine.  This combination implies you should
start the engine before you go to the front of the car'' \cite[page
237]{mellish&evans89}.

The intent of Mellish and Evans was to produce a model of a complete system as
a basis for comparison with future work.  Vander Linden's system IMAGENE can
be seen as an attempt to address one particular simplification that they made
in their work, specifically, the small range of rhetorical and grammatical
forms in the text produced by their generator.

The advantage of our system over that of Mellish and Evans, as we shall see
later, is that it is intended to be integrated into IMAGENE at some point in
the future.  However, our system produces ``flat'', linear plans rather than
hierarchical plans, so we do not need to deal with unordered actions or
abstraction hierarchies.

\subsection{Wahlster et al.}
\label{sec:wahlsteretal}

WIP, the system of Wahlster et al.\ \shortcite{wahlsteretal91,wahlsteretal93},
is a system that was designed for the generation of illustrated documents.
They argued that not only the generation of text, but also the synthesis of
multimodal documents, can be considered as a communicative act that aims to
achieve certain goals (most of which correspond to pragmatic relations in
RST).  WIP supports a plan-based approach similar to that of Moore and Paris
(see section~\ref{sec:moore&paris}); its {\it presentation planner\/} produces
a plan in the form of a directed acyclic graph, of which the leaves are
specifications for individual presentation acts, which may be realized either
in text or graphics.  The plan operators contain knowledge not only about
``what to present'' (\ie content selection), but also ``how to present'' (\ie
whether to present text or graphics); in this way, WIP interleaves content and
mode selection.  The design of WIP supports data transfer between the content
planner and the mode-specific generators, which allows for continuous
evaluation of the plan as it is produced, and revision of the initial document
structure.

The application knowledge used by WIP's presentation planner contains basic,
``compiled'' plans of the actions that need to be carried out to achieve a
task in the domain.  An example of part such a plan is the following, which
expresses that the \verb"Fill-in-water" task is achieved by carrying out the
sequence of actions \verb"Lift-lid", \verb"Remove-cover", and
\verb"Pour-water":
\begin{small}
\begin{verbatim}
      (defaction 'Fill-in-water
        (actpars ( ( ... ) )
        (sequence (A1 Lift-lid)
           (A2 Remove-cover)
           (A3 Pour-water) )
        (constraints ( ... ) )
\end{verbatim}
\end{small}
Wahlster et al.\ did not seem to have placed any emphasis on the modelling of
the domain, something which the current study investigates in some depth.

\subsection{Moore and Paris}
\label{sec:moore&paris}

Moore and Paris \shortcite{moore&paris89} constructed a text planner which is
intended to be part of an explanation facility for an expert system.  One
application which uses this planner is the Program Enhancement Advisor (PEA),
an advice-giving system which aids users in improving their Common Lisp
programs by recommending transformations that enhance the user's code.

Their planner is a top-down hierarchical expansion planner similar to that of
Sacerdoti \shortcite{sacerdoti75}, which serves to operationalize Rhetorical
Structure Theory.  The intentional, attentional, and rhetorical structure of
the generated text are recorded in the plan, as in Hovy's \shortcite{hovy88a}
planner.  The planner also makes use of a user model, which contains the
user's domain goals and assumed knowledge.

Each of their plan operators consists of the following:
\begin{description}
    \item [An effect] This is a characterisation of what goal(s) the operator
    can be used to achieve.  An effect may be an intentional goal, or a
    rhetorical relation.
    \item [A constraint list] This consists of the conditions that must be
    true for the operator to be applied, and may refer to facts in the
    system's knowledge base or in the user model.
    \item [A nucleus] This describes the main topic to be expressed.  It is
    either a primitive operator or a goal (intentional or rhetorical) that
    must be expanded further.
    \item [Satellites] These are subgoals that express additional information
    that may be needed to achieve the effect of the operator, and are
    specified as required or optional.
\end{description}

The planner works roughly as follows: when a discourse goal is posted, all the
plan operators whose effect field matches this goal are identified.  Those
operators whose constraints can be satisfied (by unification with knowledge
contained in the system's knowledge base and the user model) become candidates
for achieving the goal.  The planner chooses one on the basis of the user
model, the dialogue history, the specificity of the plan operator, and whether
or not assumptions about the user's beliefs must be made in order to satisfy
the operator's constraints.  The nucleus is then expressed.  For a primitive
goal, the corresponding text is generated; otherwise, any non-primitive
subgoals are posted for the planner to achieve recursively.  The planner
decides whether to expand optional satellites by using information from the
user model and knowledge base.

One useful consequence of this process is that the resulting (tree-shaped)
text plan contains both the intentional structure and the rhetorical structure
of the generated text.  This tree indicates which purposes different parts of
the text serve, the rhetorical means used to achieve them, and how parts of
the plan are related to each other.

Unfortunately, realisation of the primitive goals results in rather
coarse-grained text being generated.  For example, the primitive goal
\verb"(RECOMMEND S H replace-1)"
results in the following text:
\begin{quote}
    You should replace \verb"(setq x 1)" with \verb"(setf x 1)".
\end{quote}
The output of PEA at the rhetorical level, like that of WIP, is not based on
any corpus of real text.  Also, unlike IMAGENE, PEA has no provision for
expressing the leaves of its plan tree in a variety of grammatical forms.

Because PEA combines discourse knowledge with domain knowledge in its plan
operators, this knowledge is unrealistically hand-tailored to the purposes of
the planner.  It is difficult to see what use this knowledge can be put to
other than planning.  Because of the severely restricted domain of application
of such knowledge, the representations and techniques employed by systems such
as PEA leave much to be desired in view of the need for cost effectiveness
mentioned in chapter~\ref{ch:introduction}.

\subsection{Kosseim and Lapalme}
\label{sec:kosseim&lapalme}

Kosseim and Lapalme's work \shortcite{kosseim&lapalme94} focused on
determining the content and structure of instructional texts.  Their work
emphasized two types of tasks: operator tasks, \ie procedures on a system or
device to accomplish a goal external to that system/device (\eg mowing the
lawn), and maintenance/repair tasks, \ie specific operations on a
system/device
(\eg repairing a tape recorder).

Kosseim and Lapalme's system implements a two-stage process for the planning
of instructional text: a {\it task planning\/} stage, where the task
representation\footnote
    {This is a plan of the procedure, and includes a reader model and a
    domain knowledge base.}
is constructed, followed by a {\it text planning\/} stage, where the content
and rhetorical structure of the text is selected.

Task knowledge is divided into operations, preconditions, parent-child
relations\footnote
    {For example, in the sentence
    \begin{quote}
         Screw the screw-cap on the lampshade holder {\it so that you do not
         lose it}.
    \end{quote}
    which is an expression of the {\it purpose\/} rhetorical relation, the
    action is regarded as the {\it child}, and the purpose is viewed as the
    {\it parent}.},
and postconditions.  In order to map the task knowledge to the appropriate
rhetorical structure, Kosseim and Lapalme introduced an intermediate semantic
level.  This level classifies task knowledge into {\it semantic
carriers}\footnote
    {Semantic carriers represent patterns of information.  These include, for
    example, {\it sequential operation\/} (which can be expressed by the {\it
    precondition\/} or {\it action sequence\/} rhetorical relation), and {\it
    causality\/} (which can be expressed by the {\it purpose\/} or {\it
    result\/} rhetorical relation).}
according to functional criteria (the mandatory/optional nature of operations,
the execution time, the influence of an operation on the interpretation of the
procedure, etc.).  Semantic carriers help determine what task knowledge is
introduced in the text and what rhetorical relation should be used.  At the
linguistic realisation level, the actual grammatical form and position of the
rhetorical relations are selected on the basis of the results of Vander Linden
\shortcite{vanderlinden93b} adapted to French.

A corpus analysis of a wide range of operator and repair/maintenance texts was
performed.  This analysis determined:
\begin{itemize}
    \item What semantic carriers are found in the texts, where they can be
    found in the task representation, and when they are included in the texts
    (in terms of parent-child relations between nodes in the task
    representation).
    \item What rhetorical relations are used to present the semantic carriers
    and when one is preferred over another.
\end{itemize}

Kosseim and Lapalme pointed out that although there are different ways of
representing the task and interpreting the generated text, it is important
only that the reader interprets the prescribed task correctly, and that the
text seems ``natural''.  They used the notion of {\it basic-level
operations\/} introduced by Rosch \shortcite{rosch78} and Pollack
\shortcite{pollack86} for their task knowledge, on the premise that people
seem to remember and mentally represent these operations most easily.  They
remarked that these operations turn out to be detailed enough to be
descriptive, but general enough to be useful.  They also observed that
basic-level operations are a rather subjective notion and depend heavily on
factors such as the communicative goal, the discourse domain, etc.  For their
example domain of operating a VCR, their basic-level operations are:
\verb"set" any speed, \verb"select" any channel, and \verb"press" any button.
The notion of these basic-level operations is useful for helping us decide the
granularity of the actions that we need to represent in models of the device
and environment.

Kosseim and Lapalme do not pay any attention to the modelling of the
system/device, nor do they consider what knowledge is required for the
generation of warning instructions, two problems which the current study
addresses.

\section{Conclusion}

We take the stand that a complete natural language instruction generation
system for a device should have, at the top level, knowledge of the device (as
suggested by Delin et al.\ \shortcite{delinetal93}).  This is one facet of
instruction generation that the NLG systems described above (except Kosseim
and Lapalme's) have largely ignored by incorporating the {\it knowledge of the
task\/} at their top level, \ie the basic content of the instructions is
assumed to already exist and does not need to be planned for.  Kosseim and
Lapalme's system does include a task planning stage, but the knowledge used
for this planning is too superficial to be useful in generating warning
instructions.

We also believe that an NLG system that generates text of the highest quality
should use a corpus-based approach such as that of Vander Linden and Kosseim
and Lapalme, in which the rhetorical and grammatical structure of the text is
determined by features of the communicative environment, rather than an
approach such as that of Moore and Paris, in which the rhetorical structure is
determined by planning to achieve a communicative goal.  For this reason, we
wish our NLG system to perform task planning only, and leave the mapping of
the features of the instructional register to the rhetorical and grammatical
structure of the instructions (and other aspects of instructions, some of
which are discussed in section~\ref{sec:generate}) for future work.

\chapter{The knowledge base}

In this chapter, we shall examine some sample instructions and try to determine
what kind of knowledge should be stored about the artifact and the world, in
order to provide enough information for instructions to be generated.

Throughout the rest of this thesis, we shall use the term {\it
device--environment system\/} to refer to the device, the user, and any objects
or materials used by the device\footnote
     {Instances of the last in the context of the breadmaker example include
     the ingredients of the bread.}.

\section{Some example instructions and their analyses}
\label{sec:warnex}

In this section, we present several examples of instructions (taken from
\cite{breadmaker}), and analyse them to determine what types of knowledge are
necessary to understand the situations described by the sentences.  Warning
instructions have been chosen for these examples, because they serve well to
illustrate the kinds of knowledge required for instructions in general.

First, we provide some background to the breadmaker device--environment system.
In order to end up with a loaf of bread, the user should first open the lid of
the main body and remove the baking pan from the interior.  The kneading blade
should be attached to the baking pan.  Then the ingredients---the water, flour,
and yeast---should be poured, in that order, into the baking pan.  The baking
pan should then be inserted into the main body, and the {\small ON} button
pressed.  During the baking process steam will be produced in the main body as
water evaporates from the baking pan, and the steam will escape through the
steam vent.  When the breadmaker has completed the baking cycle, the baking pan
should be removed from the main body, and the bread removed from the baking
pan.

Next, we present the example instructions followed by their analyses:
\begin{enumerate}
     \item \label{en1:vent} Do not clog or close the steam vent under any
     circumstances.
     \item \label{en1:baking2} Be careful not to get burned by hot air coming
     from the machine.
     \item \label{en1:wet} Be careful not to mix the yeast with any of the wet
     ingredients (\ie water, fresh milk), otherwise, the bread may not rise
     properly.
     \item \label{en1:baking1} The main body can get very hot during the baking
     process.
     \item \label{en1:lid1} Avoid opening the lid during operation as warm air,
     which is important for proper rising, will escape.
     \item \label{en1:lid2} The lid should never be opened during the last hour
     of operation as this is the baking period.
\end{enumerate}

\subsection{Analysis of examples \protect\ref{en1:vent} and
\protect\ref{en1:baking2}}

In order to be able to reason about the situation relevant to instructions
\ref{en1:vent} and \ref{en1:baking2}, the following has to be known:

\paragraph{Steam travels through the steam vent under certain circumstances.}
It may be necessary to have a fact in world knowledge stating that in a sealed
container that has a steam vent, any steam that is produced will attempt to
escape through the steam vent.

Also, it has to be known that the circumstances under which steam is produced
can actually occur during use of the appliance.  This implies that there has to
be some kind of process description of the way in which a certain task is
carried out by the device--environment system, together with corresponding
information about {\it states} that the various components of the system go
through.

\paragraph{The steam vent is the only place through which steam can escape.}
This implies that the knowledge base must contain knowledge about the way
components of the appliance are connected to each other, that is, the relative
spatial locations of each component.  In this case, the steam vent is
``connected'' both to the inside of the container of the steam, and the outside
of the device.

\paragraph{If the steam vent is closed then steam cannot escape through it.}
It must be known that the steam vent is an {\it opening\/} to the exterior of
the device, and that any such opening could conceivably become blocked.  A
system should be able to infer the latter if it is ultimately able to produce
sentence~\ref{en1:vent}.

\paragraph{Something ``bad'' can happen if steam is not allowed to escape via
the steam vent.  The user may become burned, or the appliance may cease to
function properly.}  The possible temperature of steam must be known, and/or
the actual means by which the device could become damaged must be able to be
inferred.

\subsection{Analysis of example~\protect\ref{en1:wet}}

For instruction~\ref{en1:wet}, the following must be known:

\paragraph{The user must pour the ingredients into the baking pan at a specific
point in the task.}  This point would be defined in the process description for
making bread.

\paragraph{When one ingredient is poured on top of another, those two
ingredients become in contact with one another.}  This fact could be part of
world knowledge.

\paragraph{Yeast is involved in the rising of the bread (dough).}  This is an
example of knowledge about an object or material that is used by the device.

\paragraph{The activity of yeast may be reduced if it gets wet.}  The knowledge
base could also contain a world knowledge fact that causing a wet ingredient to
come into contact with a dry ingredient causes the dry ingredient to become wet
also.

\subsection{Analysis of example~\protect\ref{en1:baking1}}

For instruction~\ref{en1:baking1}, the following has to be known:

\paragraph{The baking process causes the breadmaker to become very hot inside,
and the heat can cause the exterior of the breadmaker to also become very hot.}
Inferring this requires knowing that during the baking process, a particular
component of the breadmaker reaches a certain temperature, and heat can be
transferred by conduction to the exterior.  This temperature may be high enough
to burn the user if he/she touches the appliance.

\subsection{Analysis of examples \protect\ref{en1:lid1} and
\protect\ref{en1:lid2}}

For sentences \ref{en1:lid1} and \ref{en1:lid2}, the following must be known:

\paragraph{The breadmaker holds warm air.  This air will escape when the lid is
open.}  This requires knowing {\it how\/} the air becomes warm, the general
fact that warm\footnote
     {The adjective {\it warm\/} in this case is used to describe the
     temperature of something relative to the outside air temperature.}
air rises, and physical knowledge of where the lid is connected in the
appliance.

\paragraph{During operation, this warm air is important for the bread to rise
properly.  If the bread does not rise properly then the final product will be
spoiled.}  This would probably be stored in the world knowledge component.  It
intuitively seems unnecessary to contemplate reasoning about the transformation
of the dough into bread at the level of molecular changes.

\section{What types of knowledge are required}
\label{sec:knowledgetypes}

The observations made in the previous section motivate our proposal that a full
knowledge base should have these components:
\begin{description}
     \item [Topological knowledge of the device] This is knowledge about the
     relative spatial locations of each component.  Some examples of
     topological knowledge are:
          \begin{itemize}
               \item \verb"handle_1" is attached to \verb"surface_1"
               \item \verb"switch_1" is located at position $x$
               \item the \verb"baking_pan" has to be at a proper orientation
               before it can click into position in the \verb"main_body"
          \end{itemize}
     \item [Kinematic knowledge of the device] This is knowledge about how the
     moving parts of the device move in relation to the other components.  For
     example:
     \begin{itemize}
          \item the \verb"washing_machine_spindle" rotates at angular velocity
          $v$ and has its axis located at position $x$
          \item the \verb"bread_slice_holder" moves in a line together with the
          \verb"start_lever"
     \end{itemize}
     \item [Electrical knowledge of the device] This should be a representation
     of the electrical circuitry of the device, possibly describing voltages,
     currents, and resistances.  This will be linked with the topological
     knowledge to some extent, in that switches, resistors, etc., are all
     physical entities that are common to both knowledge base components.
     Examples of electrical knowledge are:
          \begin{itemize}
               \item \verb"switch_1" has a resistance of 50 $\Omega$
               \item the \verb"mains_power_supply" delivers a voltage of 120 V
               across the \verb"main_circuit"
          \end{itemize}
     \item [Thermodynamic knowledge of the device] This should allow the
     specification of the {\it materials\/} comprising each component and
     physical connection.  Coupled with the electrical knowledge, the
     thermodynamic knowledge should permit the temperatures reached by each
     component to be determined, as well as the rate of increase of the
     temperatures.  For example, we could determine that:
          \begin{itemize}
               \item the \verb"heating_element" is made of tungsten
               \item the \verb"main_body" has a temperature of $T$ \dc\ at
               state $S$ of the system.  $T$ \dc\ is hot enough to burn the
               user upon contact
          \end{itemize}
     \item [Electronic knowledge of the device] This component of the knowledge
     base would only exist if the device has electronic parts.  It would
     describe the inputs and outputs of the electronic parts, possibly in the
     form of a computer program.
     \item [World knowledge] This is general knowledge that could be used in a
     variety of domains, and includes facts such as the following:
          \begin{itemize}
               \item \verb"tungsten" has a specific heat capacity of
               $t$ J K$^{-1}$ kg$^{-1}$
               \item if a dry material comes into contact with water, then the
               dry material becomes wet
               \item yeast must remain dry to be fully active
               \item a switch can be turned on by the user performing the {\it
               push\/} action on the switch
          \end{itemize}
\end{description}

\chapter{A situation calculus approach to instruction generation}
\label{ch:main}

This chapter describes one way of representing the kinds of knowledge
discussed in chapter~3, and discusses how natural language
instructions can be derived from this representation.

\section{Overview of the situation calculus}

The situation calculus (following the presentation in \cite{reiter91}) is a
first-order language that is designed to model dynamically changing worlds.  It
is based on the notion of changing {\it situations}, where the changes are the
results of a single agent performing {\it actions}.  It is assumed that the
only way in which the world can change from one state to another is by the
agent performing an action.  The initial state is denoted by the constant
$S_0$, and the result of performing an action $a$ in situation $s$ is
represented by the term $do(a,s)$.  Certain properties of the world may change
depending upon the situation.  These are called {\it fluents}, and they are
denoted by predicate symbols which take a situation term as the last argument.

An {\it action precondition axiom\/} characterizes the conditions, denoted by
$\pi_\alpha(\vec{x},s)$, under which action $\alpha(\vec{x})$ can be performed.

{\bf Action precondition axiom}
\be \label{eq:act-precond-axiom}
     Poss(\alpha(\vec{x}),s) \equiv \pi_\alpha(\vec{x},s)
\ee
For every fluent $F$, a {\it positive effect axiom\/} describes the conditions,
denoted by $\gamma^+_F(\vec{x},a,s)$, under which performing action $a$ in
situation $s$ causes $F$ to become true in the successor state $do(a,s)$.

{\bf General positive effect axiom for fluent F}
\be \label{eq:pos-effect-axiom}
     Poss(a,s) \wedge \gamma^+_F(\vec{x},a,s) \rightarrow F(\vec{x},do(a,s))
\ee
Similarly, a {\it negative effect axiom\/} describes the conditions, denoted by
$\gamma^-_F(\vec{x},a,s)$, under which performing action $a$ in situation $s$
causes $F$ to become false in the new state.

{\bf General negative effect axiom for fluent F}
\be \label{eq:neg-effect-axiom}
     Poss(a,s) \wedge \gamma^-_F(\vec{x},a,s) \rightarrow \neg
     F(\vec{x},do(a,s))
\ee
The axioms presented in this chapter have the form of
(\ref{eq:act-precond-axiom}), (\ref{eq:pos-effect-axiom}), and
(\ref{eq:neg-effect-axiom}).

Usually, {\it frame axioms\/} are also needed to specify when fluents remain
unchanged.  The {\it frame problem\/} arises because the number of frame axioms
is generally of the order of $2 \times A \times F$, where $A$ is the number of
actions and $F$ the number of fluents.

The solution to the frame problem \cite{reiter91} rests on a {\it completeness
assumption}: that the positive effect axioms describe all the ways in which
fluents can become true, and the negative effect axioms describe all the ways
in which fluents can become false.  If the completeness assumption holds, a set
of {\it successor state axioms\/} can be derived \cite{reiter91}.

{\bf Successor state axiom}
\be
     Poss(a,s) \rightarrow [F(\vec{x},do(a,s)) \equiv \gamma^+_F(\vec{x},a,s)
     \vee (F(\vec{x},s) \wedge \neg \gamma^-_F(\vec{x},a,s))]
\ee
Our current implementation uses action precondition axioms and successor state
axioms to describe the domain.

\section{Determination of the actions to be represented}
\label{sec:determineactions}

We can conceptually divide the actions that are performed in the
device--environment system into {\it reader actions\/} and {\it non-reader
actions}\footnote
     {Vander Linden \shortcite{vanderlinden93b} also makes a distinction
     between {\it reader actions\/} and {\it non-reader actions}.}.
The former are actions which can be performed by the reader of the instructions
(\ie the user of the device), whilst the latter are actions that are carried
out either by the device on its components and the materials it uses, or by
some other agent.  However, for simplicity, and because the majority of
non-reader actions are actions performed by the device, we shall only consider
device actions henceforth.

It is necessary for us to make this distinction, because natural language
instructions are directed to the user of a device, and they usually describe
mainly the actions that are executed by the user.  A device action may be
carried out by a component of the device on another component; for example, the
heating element of a toaster may carry out a {\it heating\/} action on the
bread slot, which in turn may heat the inserted bread slice.  However, we shall
not differentiate between actions performed by different components of the
system; all that need be known is that these actions are performed by the
device and not by the user.

\subsection{Ontology of high-level device actions}
\label{sec:ontology}

Device actions are the result of physical processes going on in the
device--environment system.  The device can be thought of as performing the
following high-level actions, amongst others\footnote
     {Other device actions include those related to electrical charge, but we
     do not consider those here.},
on the components and/or materials of the system:
\begin{itemize}
     \item changing the temperature of things
     \begin{itemize}
          \item heating things
          \item cooling things
     \end{itemize}
     \item moving things
     \begin{itemize}
          \item rotating things
          \item oscillating things
          \item moving things in a line/curve
     \end{itemize}
\end{itemize}
This classification should be similar to that employed by the corresponding
modules of the device design model, so that the relevant axioms (see
sections~\ref{sec:toaster-axioms} and~\ref{sec:breadmaker-axioms}) are more
easily derived.

Reasoning about the equations used to describe these physical processes is
rather complicated \cite{sandewall89,levesque&reiter95}, so instead of using
equations, we shall be using these device actions to discretely model the
continuous processes.

\section{A description of the toaster system}
\label{sec:toaster}

Table~\ref{tab:toaster-components} shows the components of the toaster and the
materials used for its operation.  Table~\ref{tab:toaster-actions} shows the
reader actions, device actions, and fluents.

\begin{table} \centering
     \begin{tabular}[t]{|l|} \hline
          {\bf Components}    \\ \hline
          {\small ON} lever   \\
          time control lever  \\
          bread slot          \\ \hline
     \end{tabular} \hspace{15mm}
     \begin{tabular}[t]{|l|} \hline
          {\bf Materials}     \\ \hline
          bread slice         \\ \hline
     \end{tabular}
     \caption{Components and materials of the toaster system}
     \label{tab:toaster-components}
\end{table}

\subsection{Meanings of the actions and fluents}

Informally, the toaster device--environment system works as follows.  The agent
(reader) can insert a slice of bread into the bread slot, and remove it from
the bread slot.  He can also press the {\small ON} lever of the toaster, which
``loads'' the bread and starts the heating process.  The act of inserting the
bread slice into the bread slot causes the bread slot to contain the bread
slice.  The bread slot ceases to contain the bread slice when the bread slice
is removed.  When the toaster pops up the bread, the bread slot is still said
to contain the (toasted) bread slice, although at this point the bread slice is
exposed (to the agent).  During the heating process, the toaster raises the
temperatures of various components and materials.

\begin{table} \centering
     \begin{tabular}[t]{|l|} \hline
     {\bf Reader actions}\\ \hline
     insert              \\
     remove              \\
     press                    \\
     touch               \\
     get\_burned         \\ \hline
     \end{tabular} \hspace{15mm}
     \begin{tabular}[t]{|l|} \hline
     {\bf Device actions}     \\ \hline
     raise\_temp              \\
     pop\_up                  \\ \hline
     \end{tabular} \hspace{15mm}
     \begin{tabular}[t]{|l|} \hline
     {\bf Fluents}  \\ \hline
     pressed        \\
     contains       \\
     removed\footnotemark\\
     temperature    \\
     touching       \\
     burned         \\
     toasted        \\
     exposed        \\ \hline
     \end{tabular}
     \caption{Reader actions, device actions, and fluents used in the toaster
     example} \label{tab:toaster-actions}
\end{table}
\footnotetext{This fluent is used only because the current implementation does
not have a representation for $\neg contains$.  We will not provide positive or
negative effect axioms for this fluent, because they are not strictly
necessary.}

\subsection{An axiomatization of the toaster system}
\label{sec:toaster-axioms}

\subsubsection{Action precondition axioms}

The following are the action precondition axioms for our toaster example.  The
domain-independent axioms are assumed to be transferable unchanged to other
domains, whereas the domain-specific axioms relate specifically to the
appliance.

When free variables appear in formulas, they are assumed to be universally
quantified from the outside.

\paragraph{Domain-independent axioms}

\beq \label{eq:pre-insert}
     Poss(insert(x,y),s) \equiv
          three\_d\_location(y) \wedge fits(x,y) \wedge exposed(y,s)
\eeq\beq \label{eq:pre-remove}
     Poss(remove(x,y),s) \equiv
          three\_d\_location(y) \wedge contains(y,x,s) \wedge exposed(x,s)
\eeq\beq \label{eq:pre-press}
     Poss(press(x),s) \equiv button(x) \vee lever(x)
\eeq\beq \label{eq:pre-touch}
     Poss(touch(x),s) \equiv
          physical\_object(x) \wedge exposed(x,s)
\eeq\beq \label{eq:pre-getburned}
     Poss(get\_burned,s) \equiv
          \exists x,t.(touching(x,s) \wedge temperature(x,t,s) \wedge t \geq
          70)
\eeq

Axiom~(\ref{eq:pre-insert}) states that an action by the agent of inserting $x$
into $y$ is possible in state $s$ if $y$ is a {\it three\_d\_location}, \ie a
spatial volume, $x$ fits into $y$, and $y$ is exposed.  Note that this axiom
attempts to capture only one sense of the meaning of {\it insert}.

Axiom~(\ref{eq:pre-remove}) expresses that an action of removing $x$ from $y$
is possible in state $s$ if $y$ is a {\it three\_d\_location}, $x$ is contained
in $y$, and $x$ is exposed, in state $s$.

Axiom~(\ref{eq:pre-press}) asserts that the {\it press\/} action is only
possible on buttons or levers.  This is clearly an incomplete formalization if
we wish to describe the broad meaning of {\it press\/} (\eg a surface can also
be pressed), but it is sufficient for our purposes.

Axiom~(\ref{eq:pre-touch}) states that it is possible for the agent to touch an
object if it is exposed.  The fluent {\it exposed\/} is useful for describing
the conditions under which harm can occur to the agent, as we shall see in more
detail in section~\ref{sec:derivewarning}.

Axiom~(\ref{eq:pre-getburned}) asserts that an action of the agent getting
burned is possible if the agent is touching something with a temperature of at
least 70 \dc.

The rest of the axioms are straightforward.

\paragraph{Domain-specific axioms}

\beq \label{eq:pre-raisetemp}
     \lefteqn{Poss(raise\_temp(x),s) \equiv (x=bread\_slot \vee
     contains(bread\_slot,x,s)) \wedge} \nl
          & & \exists t.(temperature(x,t,s) \wedge t<200) \wedge
          pressed(on\_lever,s)
\eeq\beq \label{eq:pre-popup}
     Poss(pop\_up,s) \equiv
          \exists t.(temperature(bread\_slot,t,s) \wedge t \geq 200)
\eeq

Axiom~(\ref{eq:pre-raisetemp}) states that an action by the device of raising
the temperature of component (or material) $x$ is possible only if the
temperature of $x$ is less than 200 \dc\footnote
     {We make an assumption here that the electrical subsystem constrains the
     maximum temperature of any component.},
and the {\small ON} lever is pressed.  This axiom currently considers only the
bread slot and bread slice to be components and materials of the toaster system
(see section~\ref{sec:deriveaxioms} for a discussion on this).

Axiom~(\ref{eq:pre-popup}) states that the device can cause the bread slot to
pop up its contents if the temperature of the bread slot reaches 200 \dc.  This
is a temporary simplification of stating that the popping up action is possible
if the temperature of the bread slot remains 200 \dc\ for some period of time.

For our simple toaster example, it is merely coincidental that the reader
actions are all domain-independent, and the device actions are all
domain-specific.  A more complex device may allow an action upon it that is
peculiar to that device; also, a host of devices may share common actions on
their components or materials, such as spinning, etc.

\subsubsection{Positive effect axioms}

\paragraph{Domain-independent axioms}

\beq \label{eq:pos-contains}
     Poss(a,s) \wedge a=insert(x,y)
          \rightarrow contains(y,x,do(a,s))
\eeq\beq \label{eq:pos-pressed}
     Poss(a,s) \wedge a=press(x)
          \rightarrow pressed(x,do(a,s))
\eeq\beq \label{eq:pos-touching}
     Poss(a,s) \wedge a=touch(x)
          \rightarrow touching(x,do(a,s))
\eeq\beq \label{eq:pos-burned}
     Poss(a,s) \wedge a=get\_burned
          \rightarrow burned(do(a,s))
\eeq

Axiom~(\ref{eq:pos-contains}) asserts that inserting $x$ into $y$ in state $s$
results in $y$ containing $x$ in state $do(a,s)$.

Axiom~(\ref{eq:pos-pressed}) states that the action of pressing $x$ in state
$s$, provided it is possible, results in $x$ becoming {\it pressed\/} in state
$do(a,s)$.

Axiom~(\ref{eq:pos-touching}) states that a {\it touch\/} action (by the agent)
on $x$ results in the agent touching $x$ in the new state.

Axiom~(\ref{eq:pos-burned}) states that if it is possible for the agent to get
burned (by the {\it get\_burned\/} action), then the agent may be burned in the
new state.

\paragraph{Domain-specific axioms}

\beq \label{eq:pos-exposed}
     Poss(a,s) \wedge a=pop\_up \wedge contains(bread\_slot,x,s)
          \rightarrow exposed(x,do(a,s))
\eeq\beq \label{eq:pos-temperature1}
     \lefteqn{Poss(a,s) \wedge a=raise\_temp(x) \wedge temperature(x,t,s)
     \rightarrow} \nl
          & & temperature(x,t+50,do(a,s))
\eeq\beq \label{eq:pos-temperature2}
     Poss(a,s) \wedge a=pop\_up
          \rightarrow temperature(x,20,do(a,s))
\eeq

Axiom~(\ref{eq:pos-exposed}) expresses that if the device causes $x$ to pop up
in state $s$, then $x$ becomes exposed in the next state.

Axiom~(\ref{eq:pos-temperature1}) states that the {\it raise\_temp\/} action on
component (or material) $x$ causes the temperature of $x$ to increase by 50
\dc\ in the successor state.

As a simplification of the fact that all the components of the toaster system
eventually cool down to room temperature after the {\it pop\_up\/} action,
axiom~(\ref{eq:pos-temperature2}) states that their temperatures become equal
to room temperature\footnote
     {Assumed to be a constant at 20 \dc.}
instantaneously.

\subsubsection{Negative effect axioms}

\paragraph{Domain-independent axioms}

\beq \label{eq:neg-contains}
     Poss(a,s) \wedge a=remove(x,y)
          \rightarrow \neg contains(y,x,do(a,s))
\eeq

\paragraph{Domain-specific axioms}

\beq \label{eq:neg-pressed}
     Poss(a,s) \wedge a=pop\_up
          \rightarrow \neg pressed(on\_lever,do(a,s))
\eeq\beq \label{eq:neg-exposed}
     \lefteqn{Poss(a,s) \wedge a=press(on\_lever) \wedge
     contains(bread\_slot,x) \rightarrow} \nl
          & & \neg exposed(x,do(a,s))
\eeq

Axiom~(\ref{eq:neg-pressed}) expresses that the {\it pop\_up\/} action results
in the {\small ON} lever no longer being pressed; this is a mechanical action
by the device, and indicates the end of the toasting process.

Axiom~(\ref{eq:neg-exposed}) states that an action of the reader pressing the
{\small ON} lever causes anything in the bread slot to become unexposed; this
happens because the object in the bread slot gets ``pushed down''.  As we have
already seen, the related positive effect axiom~(\ref{eq:pos-exposed}) causes
this object to become exposed once again when the {\it pop\_up\/} action is
performed.

\section{Deriving instructions from the axioms}
\label{sec:deriveinstructions}

As in \cite{pinto94}, we shall abbreviate terms of the form:
\[ do(a_n,(do(\ldots,do(a_1,s)\ldots)) \]
as:
\[ do([a_1,\ldots,a_n],s). \]

Our aim is to derive a sequence of actions (reader and device) which, when
performed, causes a slice of bread to become toasted.  Ideally, this sequence
would begin with the act of the reader inserting a fresh slice of bread into
the toaster, and end with the act of the reader removing the toasted bread from
the toaster.  A typical sequence of reader actions could be as follows:
\begin{enumerate}
     \item Adjust time lever for desired degree of toasting.
     \item Insert a slice of bread into the bread slot.
     \item Press the {\small ON} lever.
     \item Remove the toast when it pops up.
\end{enumerate}
The device actions are interleaved in some fashion with the reader actions, but
many sequences of instructions do not refer to device actions.

We need to formulate the goals for the planner in order to be able to come up
with something resembling the above sequence.  Since the overall goal of the
reader in using the toaster is to toast a slice of bread, it makes sense to
describe the goal in terms of the final state of the material (bread, in this
case)\footnote
     {The goals of an agent in using other kitchen appliances can be expressed
     in terms of the final states of the materials they operate upon: a washing
     machine delivers clean clothes, a kettle produces hot water, a breadmaker
     produces bread, etc.}.
The plan will then describe a sequence of device and reader actions which cause
the transformation of the material from its initial to its desired state.  Note
that we make a distinction between the states of individual components and
materials of the device--environment system, and the global state of the whole
system.

Let us see what happens if we chose the goal of the system to be to produce a
piece of toast.  How do we formulate this goal?  First of all, we need fluents
to describe the states of all the components and materials.  As a reasonable
approximation, we could model the state changes of the bread in terms of the
temperature of the bread.  Using $temperature(x,t,s)$ as a fluent describing
that object $x$ has a temperature of $t$ \dc\ in state $s$, we could simply
define toast as a slice of bread that has reached a temperature of 200 \dc:
\beq \label{eq:toasted}
     \lefteqn{toasted(bread\_slice,do(a,s)) \leftarrow} \nl
          & & temperature(bread\_slice,t,s) \wedge t \geq 200 \vee
          toasted(bread\_slice,s)
\eeq
Note that using this definition, {\it toasted(bread\_slice)\/} holds for all
states after {\it do(a,s)}.  So, the bread slice remains toasted even when its
temperature falls below 200 \dc.

Let $S_0$ denote the initial global state, in which the bread slot and bread
slice are both at room temperature (20 \dc), and the bread slot and bread slice
are exposed (\ie the agent can touch them).  Figure~\ref{fig:initstate} shows
the fluents that hold in this initial state.

\begin{figure} \centering
     \begin{tabular}{l}
          $temperature(bread\_slot,20,S_0)$ \\
          $temperature(bread\_slice,20,S_0)$ \\
          $exposed(bread\_slot,20,S_0)$ \\
          $exposed(bread\_slice,20,S_0)$
     \end{tabular}
     \caption{Fluents that hold in the initial state, $S_0$}
     \label{fig:initstate}
\end{figure}

Then, a possible plan to cause this fluent to become true could be this:
\beq
     \lefteqn{do([insert(bread\_slice,bread\_slot), press(on\_lever),
     raise\_temp(bread\_slice),} \nl
     & & raise\_temp(bread\_slice), raise\_temp(bread\_slice),
     raise\_temp(bread\_slice)],S_0)
\eeq
The {\it raise\_temp\/} action is carried out four times, since each time it
raises the temperature of something by 50 \dc.  This sequence of actions does
cause the slice of bread to become toasted, but it does not say anything about
finishing off the process that instructions usually talk about; that is,
causing the slice to pop up and having the reader remove it.  This can easily
be accomplished by adding an extra condition to the goal $G$\footnote
     {Note that $G$ should consist of fluent expressions which by definition
     must contain a state variable, but this expression contains terms which
     lack state variables.  We can think of these terms as {\it representing\/}
     true fluent expressions.  When used for reasoning, the state variables are
     restored.}:
\be
     G = toasted(bread\_slice) \wedge removed(bread\_slice,bread\_slot)
\ee
A possible plan then becomes this:
\beq
     \lefteqn{do([insert(bread\_slice,bread\_slot), press(on\_lever),
     raise\_temp(bread\_slice),} \nl
     & & raise\_temp(bread\_slice), raise\_temp(bread\_slice),
     raise\_temp(bread\_slice), \nl
     & & pop\_up, remove(bread\_slice,bread\_slot)],S_0)
\eeq
The instruction sequence corresponding to this plan could be this:
\begin{enumerate}
     \item Insert the bread slice into the bread slot.
     \item Press the {\small ON} lever.
     \item When the bread slice pops up, remove it from the bread slot.
\end{enumerate}
Note that this sequence does not include any references to the time control
lever: this lever determines the length of time that the bread slice will be
heated for, and we have not included any knowledge of this in our
axiomatization.

Also note that we do not model the perception actions of the reader watching
for the bread slice to pop up.  In our simple domain, we have avoided the need
for these by assuming that the reader knows when a salient observable change
occurs in the system.  In this case, the salient change is the popping up of
the bread slice.

\section{Deriving warning instructions}
\label{sec:derivewarning}

Many instructional texts contain warning and safety instructions mingled, or
together with, the basic procedural instructions.  In order for us to generate
warning instructions we must be able to derive possible plans, using the
available actions and fluents, in which the reader can become harmed.  There
are many ways in which this can happen: by burning, electric shock, laceration,
crushing, etc.  For each different type of injury, different factors need to be
considered.  So, for example, when reasoning about the possibility of an
electric shock, the electrical subsystem and related components must be
examined; for the possibility of laceration, sharp objects must be considered.
We shall concentrate on examining the conditions under which burns to the user
can occur.  For this, we must consider thermal properties of the objects in the
device--environment system.  As a crude approximation to the modelling of
thermodynamics in our system, we shall only regard the absolute temperature of
the objects to be significant.  These values can be derived from lower-level
physical knowledge such as the topology and heat conductivity of the various
components, and knowledge of the electrical subsystem (see
\cite{sandewall89,levesque&reiter95} for suggestions for the modelling of
continuous, physical processes).

We can derive a plan for which the user gets burned by setting the goal $G$ to
be this:
\be
     G = burned
\ee
A possible plan would then be this:
\beq
     \lefteqn{do([insert(bread\_slice,bread\_slot), press(on\_lever),
     raise\_temp(bread\_slot),} \nl
     & & raise\_temp(bread\_slice), raise\_temp(bread\_slot), \nl
     & & touch(bread\_slot), get\_burned],S_0)
\eeq

It is clear that the penultimate action in this plan is the one which causes
the agent to become burned, as can be seen from axiom~(\ref{eq:pos-touching});
the previous actions are those that make this {\it touch\/} action possible.
Hence, the appropriate warning instruction should be something like this:
\be \label{sen:notouchslot}
     \mbox{Do not touch the bread slot during the heating period.}
\ee

We now have two problems.  Firstly, we need to determine where this caution
should be placed in the instruction sequence.  Secondly, we need to be able to
refer to a sequence of similar actions by a generic name; in this case, the two
{\it raise\_temp\/} actions are collectively called the {\it heating period}.

\subsection{Determining the placement of warning instructions}
\label{sec:placement}

For this problem, a solution would be to add the fluent {\it burned\/} to the
goal, so that:
\be
     G = toasted(bread\_slice) \wedge removed(bread\_slice,bread\_slot) \wedge
     burned
\ee
Planning would continue as normal, with the {\it get\_burned\/} action being
included in the plan.  At the point in the sentence plan where the {\it
get\_burned\/} action is encountered, a negative imperative caution, such as
sentence~(\ref{sen:notouchslot}), will be generated.  The goal of the agent
being burned should not be thought of as having been achieved: the {\it
get\_burned\/} action merely denotes a point where the potential for injury
exists.

Following this approach, one possible plan which includes the {\it
get\_burned\/} action is this:
\beq
     \lefteqn{do([insert(bread\_slice,bread\_slot), press(on\_lever),
     raise\_temp(bread\_slot),} \nl
     & & raise\_temp(bread\_slice), raise\_temp(bread\_slot), \nl
     & & touch(bread\_slot), get\_burned, raise\_temp(bread\_slice), \nl
     & & raise\_temp(bread\_slot), raise\_temp(bread\_slice), \nl
     & & raise\_temp(bread\_slot), raise\_temp(bread\_slice), \nl
     & & pop\_up, remove(bread\_slice,bread\_slot)],S_0)
\eeq

We can imagine processes in which there are many possible situations where the
user can get hurt.  Since the potential for being burned may exist in more than
one situation, once the {\it get\_burned\/} action is planned for, the goal
{\it burned\/} should not be discarded: following Hovy \shortcite{hovy88a}, we
consider that such a goal needs to be planned for {\it in-line}.  In this
approach, the planner completes its task by planning in-line, during
realization.  For our purposes, this means that after the basic plan is
obtained, the plan is examined for places in which the {\it touch\/} and {\it
get\_burned\/} actions (together) could be inserted, \ie places where the {\it
get\_burned\/} action can be planned for using only reader actions.  This
simply requires checking all the places in the plan where these actions'
preconditions are satisfied.

This technique does pose a problem, however.  If the planner were allowed to
insert {\it touch\/} and {\it get\_burned\/} actions wherever possible, the
resulting plan could be something like this:
\beq \label{eq:toomanyinjuries}
     \lefteqn{do([[insert(bread\_slice,bread\_slot)], [press(on\_lever)],} \nl
     & & [raise\_temp(bread\_slot)], [raise\_temp(bread\_slice)], \nl
     & & [raise\_temp(bread\_slot), touch(bread\_slot), get\_burned], \nl
     & & [raise\_temp(bread\_slice), touch(bread\_slot), get\_burned], \nl
     & & [raise\_temp(bread\_slot), touch(bread\_slot), get\_burned], \nl
     & & \vdots \nl
     & & [pop\_up], [remove(bread\_slice,bread\_slot)]],S_0)
\eeq
In this plan, extra square brackets have been added to illustrate the grouping
of actions which results after in-line planning has been performed.

This problem can be solved by first finding a solution to the other problem of
generating sentence~(\ref{sen:notouchslot}), that of being able to refer to a
set of similar actions collectively.  Then, we could simply constrain the
planner to attempt to plan for one injury (for each injury type) per
collection.

Notice that we have made some important simplifications here.  Realistically,
the sequence of actions leading to one particular type of injury in a time
period may not be unique.  There may be many ways of achieving the injury;
indeed, given a more complex model of reader interactions with the
device--environment system, there may be infinitely many such sequences.  An
implementation taking this into account should therefore place a bound on the
length of the injury sequences planned for, and it should incorporate
heuristics indicating which sequences are too unlikely to warrant
consideration.  For example, an injury sequence of, say, four or five reader
actions might be ignored because it is highly unlikely that the reader would
carry out such a sequence.

Also, the actions in each injury sequence need not necessarily all be reader
actions.  For our simple approach, this requirement is justifiable because
there is only one possible action---the {\it touch\/} action---that can cause
an injury.  If device actions were allowed, then the planner could insert {\it
several of the following actions of the basic plan}, as well as the extra
actions leading to the injury, into many points of the basic plan, as in the
following:
\beq
     \lefteqn{do([[insert(bread\_slice,bread\_slot)], [press(on\_lever),} \nl
     & & raise\_temp(bread\_slot), raise\_temp(bread\_slice), \nl
     & & raise\_temp(bread\_slot), touch(bread\_slot), get\_burned], \nl
     & & \vdots \nl
     & & [pop\_up], [remove(bread\_slice,bread\_slot)]],S_0)
\eeq
The injury sequence in this plan is clearly undesirable, because it includes
several actions of the basic plan.  Therefore, some technique would have to be
implemented that disallows more than one normal action of the basic plan to be
included in the injury sub-plan.  One such technique would be for the planner,
when performing in-line planning, to not consider sub-plans where the next
action is identical to the next action in the basic plan.

\subsection{Collecting similar actions together}
\label{sec:collect}

Hovy \shortcite{hovy88a} gives an example of a straightforward text generated
by his PAULINE system:
\begin{quote}
     First, Jim bumped Mike once, hurting him.  Then Mike hit Jim, hurting him.
     Then Jim hit Mike once, knocking him down.  Then Mike hit Jim several
     times, knocking him down.  Then Jim slapped Mike several times, hurting
     him.  Then Mike stabbed Jim.  As a result, Jim died.
\end{quote}
By grouping together similar enough topics, and then generating the groupings
instead of the individual actions, we can formulate less tedious variants of
the text.  For this example, using the force of the actions as the similarily
criterion, PAULINE can produce the following variants:
\begin{enumerate}
     \item \label{sen:fight1} Jim died in a fight with Mike.
     \item \label{sen:fight2} After Jim bumped Mike once, they fought, and
     eventually Mike killed Jim.
     \item \label{sen:fight3} After Jim bumped Mike once, they fought, and
     eventually he was knocked to the ground by Mike.  He slapped Mike a few
     times. Then Mike stabbed Jim and Jim died.
\end{enumerate}
In variant~(\ref{sen:fight1}), all actions were grouped together; in
variant~(\ref{sen:fight2}), all actions more violent than bumping but less
violent than killing were accepted; and in variant~(\ref{sen:fight3}), the
grouping resulted from defining four levels of violence: bumping, hitting and
slapping, knocking to the ground, and killing.

Hovy asserts that this technique of grouping together actions by levels of
force is very specific and not very useful.  He gives examples of
generator-directed inference, such as recognizing that in a political
nomination race one candidate {\it beats\/} another candidate, because both
voting outcomes relate to one election and the winning candidate has a higher
number of votes than the losing one.  This is termed an {\it interpretation},
because a new concept {\it beat\/} is formed by interpreting the input as an
instance of {\it beat}.

In section~\ref{sec:ontology} we explained our rationale for attempting to
approximate continuous physical processes by the use of discrete states and
device actions.  For our toaster example, the continuous process of a slice of
bread being heated was represented by a series of {\it raise\_temp\/} actions
by the toaster.  Thus, a process which might be described using just one
equation is represented by a series of contiguous, repeated actions.  The
continuous process might be assigned a name, such as ``the heating period'' or
``the kneading stage'', so we want to be able to refer to the corresponding
series of actions by such names, \ie we want to refer to the {\it
interpretations\/} rather than the {\it details}.

Note that the action plan may contain interleaved device actions as a result of
the approximation of a continuous process involving more than one component or
material: the heating of a slice of bread involves raising the temperature of
the bread itself, the bread slot, and the heating element.  In reality these
temperatures rise in tandem with one another (though these temperatures won't
be equal), but the planner will interleave the actions.  Since all these
heating actions are closely related to one another, we need some provision for
establishing that they together constitute an overall heating period.  A
simplified way of determining this is to simply recognize that these actions
are grouped together in the plan.  A more complicated approach would be to also
consider the spatial and thermodynamic relationships between the components
affected by these actions: we can imagine a situation in which a large system
has two ``distant'' components which just happen to be getting heated
concurrently, but we cannot refer to the heating periods using a common term
such as ``heating period''.  We shall adopt the former, simple,
approach.\footnote
     {Observe that if we used a continuous, rather than discrete, approach such
     as that of Levesque and Reiter \shortcite{levesque&reiter95}, then the
     need to group actions like this would be avoided.}

The inferred concepts referred to by instructional texts are generally very
simple.  For the toaster, the only interpretation that needs to be made is that
the sequence of {\it raise\_temp\/} actions forms a period which we might call
``the heating period''.  One question which immediately arises is, how many
times does an action have to be repeated to merit a label being ascribed to the
sequence?  The answer depends on several factors.  A label will need to be
assigned if:
\begin{enumerate}
     \item The process represented by the sequence of actions is referred to
     anywhere in the generated text.  For simplicity and flexibility, we shall
     assume that a label always needs to be inferred.
     \item The granularity of the actions is large, or the number of
     repetitions is high and the granularity is sufficient (indicating that the
     continuous process was taking place over a significant length of time).
     The granularity of the actions will have been determined during the
     derivation of the axioms.
\end{enumerate}

We shall need a place to store the results of the interpretations, so that the
final stage, the {\it instruction realization\/} stage, can use them.  Since an
inferred label may span several actions, it is useful to {\it index\/} the
actions in non-descending order, such that all those actions belonging to one
collection have the same index.  This method allows our system to easily
inspect whether an action takes place {\it during\/} the period represented by
a label.  The labels, together with their corresponding indices, will simply be
stored in a separate list.

After all the possible sequences leading to an injury have been planned for
(giving us sequence~(\ref{eq:toomanyinjuries})), the interpretations have been
performed, and the superfluous injury sequences subsequently removed, the final
sequence of indexed actions is the following:
\beq \label{eq:indexedactions}
     \lefteqn{do([insert(bread\_slice,bread\_slot)^{(1)},
     press(on\_lever)^{(2)},  raise\_temp(bread\_slot)^{(3)},} \nl
     & & raise\_temp(bread\_slice)^{(3)}, raise\_temp(bread\_slot)^{(3)}, \nl
     & & touch(bread\_slot)^{(3)}, get\_burned^{(3)},
     raise\_temp(bread\_slice)^{(3)}, \nl
     & & raise\_temp(bread\_slot)^{(3)}, raise\_temp(bread\_slice)^{(3)}, \nl
     & & raise\_temp(bread\_slot)^{(3)}, raise\_temp(bread\_slice)^{(3)}, \nl
     & & pop\_up^{(4)}, remove(bread\_slice,bread\_slot)^{(5)}],S_0)
\eeq
and the label list just contains the information that index (3) refers to the
heating period:
\be
     [(3,heating\_period)]
\ee

There are also situations in which interpretations should be performed that do
not involve continuous physical processes.  Consider the following subsequence
of instructions, taken from the breadmaker domain (see
section~\ref{sec:breadmaker}):
\begin{enumerate}
     \item Pour the water into the baking pan.
     \item Pour the flour into the baking pan.
     \item Pour the yeast into the baking pan.
\end{enumerate}
If the order of these actions were not important, then this subsequence could
be replaced by the following single instruction:
\be
     \mbox{Pour the ingredients into the baking pan.}
\ee
However, deciding the relative importance of reader actions is beyond the scope
of this work, so in our system this particular type of interpretation will not
be performed.

\section{Generating the instructions}
\label{sec:generate}

We shall use the Penman system (see section~\ref{sec:penman}) to generate the
instructions.  Penman's inputs are primarily organized around {\it processes},
which include actions, events, states, relations, etc.

An action, event, or state contains some number of entities that participate in
the actualization of that action, event, or state.  The manner of these
entities' participation is identified in terms of given role names.

In order to make Penman generate a sentence, the process that the sentence is
based upon must be specified.  We shall be using SPL (Sentence Plan Language)
in order to specify the actions, together with the roles of the actions and
their fillers.

\subsection{Determining the role fillers}
\label{sec:determineroles}

Each argument position of an action is designated exactly one role, the filler
of that position being the filler for that role.  The agent of the action is
determined by whether that action is a {\it reader action\/} or a {\it device
action}.  So, for example, the action $insert(bread\_slice,bread\_slot)$
describes an {\it insert\/} action with the agent as the {\small ACTOR}, the
{\it bread\_slice\/} as the {\small ACTEE}, and the {\it bread\_slot\/} as the
{\small DESTINATION}.  Table~\ref{tab:designatedroles} lists the possible
actions in our system, together with the roles and arguments associated with
their arguments.

\begin{table} \centering
     \begin{tabular}{|l|l|l|} \hline
     {\bf Action}        & {\bf Role}   & {\bf Filler} \\ \hline
     $insert(x,y)$       & {\small ACTOR}         & reader       \\
                         & {\small ACTEE}         & $x$          \\
                         & {\small DESTINATION}   & $y$          \\ \hline
     $remove(x,y)$       & {\small ACTOR}         & reader       \\
                         & {\small ACTEE}         & $x$          \\
                         & {\small SOURCE}        & $y$          \\ \hline
     $press(x)$               & {\small ACTOR}         & reader       \\
                         & {\small ACTEE}         & $x$          \\ \hline
     $touch(x)$          & {\small ACTOR}         & reader       \\
                         & {\small ACTEE}         & $x$          \\
     $get\_burned$       & {\small ACTOR}         & reader       \\ \hline
     $raise\_temp(x)$    & {\small ACTOR}         & device       \\
                         & {\small ACTEE}         & $x$          \\ \hline
     $pop\_up$           & {\small ACTOR}         & device       \\ \hline
     \end{tabular}
     \caption{Roles of actions} \label{tab:designatedroles}
\end{table}

The information gathered by the interpretation stage, as described in
section~\ref{sec:collect}, gives us the {\small EXHAUSTIVE-DURATION}
role\footnote
     {This, in Penman terminology, means the time period during which an
     action, event, or process occurs.}
of the actions which take place during that period.

Table~\ref{tab:defaultroles} shows the basic roles and fillers of actions that
are typically assumed by instructional texts.

For the {\it touch\/} action the polarity role is assigned a filler of
``negative'' when it is part of a sequence ending in an injury to the user.
This is because the {\it touch\/} action should not be performed by the agent
under the circumstances; this filler overrides the default filler of
``positive''.  Note that, as we discussed in section~\ref{sec:placement}, the
{\it touch\/} action is used for planning injury sequences.  In our simple
model, this action can occur {\it only\/} in injury sub-plans, so we can
automatically assume that it should generate a warning instruction.  However,
in a more complex model, reader actions may occur in the basic plan, and in
this case, any reader actions that should generate warnings should be specially
tagged in the plan representation.

Vander Linden \shortcite{vanderlinden93b} focuses on determining the features
that contribute to variation in natural language instructions; this is beyond
the scope of the current work.

\begin{table} \centering
     \begin{tabular}{|l|l|} \hline
          {\bf Role}          & {\bf Filler} \\ \hline
          {\small TENSE}      & present      \\
          {\small SPEECHACT}  & imperative   \\
          {\small POLARITY}   & positive     \\ \hline
     \end{tabular}
     \caption{Default roles of actions in non-warning instructions}
     \label{tab:defaultroles}
\end{table}

\subsection{Mechanisms for determining some other role fillers}

The mechanism described in section~\ref{sec:collect} for grouping together
similar actions allows us to determine only the {\small EXHAUSTIVE-DURATION}
role of the actions that took place during that period.

Sometimes we would like the system to infer the fillers of other roles for a
particular action.  For example, sometimes we will want the system to generate
an {\it explanation\/} instruction (see section~\ref{sec:paris&scott}), in
other words, an instruction containing both a {\it matrix\/} and a {\it
purpose\/} clause \cite{dieugenio92b}.  An instance of this would be the
following warning instruction:
\be
     \mbox{Do not touch the bread slot during the heating period to avoid
     getting burned.}
\ee
In our representation, determining the {\it range\/} of the rhetorical relation
{\small RST-PURPOSE} (\ie the purpose clause) simply amounts to following the
chain of actions in the plan, beginning with the {\it touch\/} action, to the
next injury action.  Nevertheless, determining the conditions under which a
purpose clause should be generated is beyond the scope of this work (but see
\cite{vanderlinden93b}), so we shall not consider this issue further.  By
default, our system will only generate matrix clauses and will omit purpose
clauses.

\subsection{Deciding which actions to mention}
\label{sec:otheractions}

As we have noted previously (see section~\ref{sec:determineactions}),
instruction sequences describe mainly actions that are carried out by the user
of the device.  So, we may assume that our system should generate instructions
corresponding to these reader actions.  However, there are situations in which
it is appropriate to mention device actions.  Consider, for example, the
following sentences:
\begin{enumerate}
     \item The bread slice will pop up.
     \item Remove it from the bread slot.
\end{enumerate}
Intuitively, the popping up action is mentioned because the bread slice had
been contained in the bread slot for some period of time during which it was
concealed from the agent, or ``unexposed''.  Since the popping up action was
the main event which indicated the completion of the toasting process to the
agent (because this action caused the bread slice to suddenly become exposed
again), we deem this device action important enough to mention.  Hence, we can
state this intuition more generally by saying that if there is a significant
period of time during which the agent is not involved in the process, when the
agent {\it should\/} eventually perform an action, then the last conspicuous
device action during that period should be mentioned.  In our representation
formalism, this can be stated thus:
\begin{itemize}
     \item If the interpretation stage inferred a collection of device actions,
     this collection is followed by reader action $A_{reader}$, and the last
     device action $A_{device}$ caused a change of state of a salient exposed
     component or material\footnote
          {For our toaster example, the bread slice, the principle material
          operated upon, is the one that should be considered.},
     then construct SPL specifications for both actions $A_{device}$ and
     $A_{reader}$.
\end{itemize}

We do not propose a principled approach for selecting which actions should be
mentioned because there are many factors that contribute towards these
decisions; an accurate assessment of these factors could be made by a study
essentially similar to that of Vander Linden's \shortcite{vanderlinden93b}.
Instead of attempting to characterize these features, we shall just simply
abstract them in the form of the pattern presented above.

Note also that not {\it every\/} reader action will be mentioned in the
instructions.  Consider the following subsequence of instructions:
\begin{enumerate}
     \item The ``complete'' light will flash when bread is done.
     \item Remove baking pan from unit.
\end{enumerate}
In between these two actions, the user of the device must open the lid.  This
is considered too ``obvious'' to mention, possibly because something similar
was mentioned elsewhere in the instructions.  For simplicity, our system
mentions every reader action.

\subsection{A sample generated instruction sequence}

After the role fillers of the actions (as in
sequence~(\ref{eq:indexedactions})) have been determined using the types of the
actions and their arguments, the consequent SPL specification would result in
Penman generating the following natural language instruction sequence (see
appendix~\ref{app:a} for the complete output of the system):
\begin{small}
\begin{verbatim}
     Insert the bread slice into the toaster's bread slot.
     Press the ON lever.
     Do not touch the toaster's bread slot during the heating period.
     The bread slice will pop up.
     Take the bread slice out of the toaster's bread slot.
\end{verbatim}
\end{small}

\section{Another example: the breadmaker system}
\label{sec:breadmaker}

Table~\ref{tab:breadmaker-components} shows the components of the breadmaker
and the materials used for its operation.  Table~\ref{tab:breadmaker-actions}
shows the reader actions, device actions, and fluents used in our breadmaker
example.

\begin{table} \centering
     \begin{tabular}[t]{|l|} \hline
          {\bf Components}    \\ \hline
          {\small ON} button  \\
          main body           \\
          main body interior  \\
          baking pan          \\
          baking pan interior \\
          kneading blade      \\
          lid                 \\
          steam vent          \\
          ``complete'' light  \\ \hline
     \end{tabular} \hspace{15mm}
     \begin{tabular}[t]{|l|} \hline
          {\bf Materials}     \\ \hline
          water               \\
          flour               \\
          yeast               \\
          bread               \\ \hline
     \end{tabular}
     \caption{Components and materials of the breadmaker system}
     \label{tab:breadmaker-components}
\end{table}

\subsection{Meanings of the actions and fluents}

Informally, the breadmaker device--environment system works as follows.  In
order to end up with a loaf of bread, the user should first open the lid of the
main body and remove the baking pan from the interior.  The kneading blade
should be attached to the baking pan.  Then the ingredients---the water, flour,
and yeast---should be poured, in that order, into the baking pan.  The baking
pan should then be inserted into the main body, and the {\small ON} button
pressed.  During the baking process the main body will become ``steamified'',
\ie steam will be produced there, and the steam will escape through the steam
vent.  When the breadmaker has completed the baking cycle, the ``complete''
light will flash.  The baking pan should then be removed from the main body,
and the bread removed from the baking pan.  During the heating process, the
breadmaker raises the temperatures of various components and materials.

\begin{table} \centering
     \begin{tabular}[t]{|l|} \hline
     {\bf Reader actions}\\ \hline
     insert              \\
     attach              \\
     pour                \\
     remove              \\
     press                    \\
     open                \\
     close               \\
     touch               \\
     get\_burned         \\ \hline
     \end{tabular} \hspace{15mm}
     \begin{tabular}[t]{|l|} \hline
     {\bf Device actions}     \\ \hline
     raise\_temp              \\
     steamify                 \\
     flash                    \\ \hline
     \end{tabular} \hspace{15mm}
     \begin{tabular}[t]{|l|} \hline
     {\bf Fluents}  \\ \hline
     pressed        \\
     contains       \\
     attached       \\
     removed        \\
     opened         \\
     flashing       \\
     temperature    \\
     touching       \\
     burned         \\
     exposed        \\ \hline
     \end{tabular}
     \caption{Reader actions, device actions, and fluents used in the
     breadmaker example} \label{tab:breadmaker-actions}
\end{table}

\subsection{An axiomatization of the breadmaker system}
\label{sec:breadmaker-axioms}

\subsubsection{Action precondition axioms}

\paragraph{Domain-independent axioms}

\beq \label{eq:pre-insertB}
     Poss(insert(x,y),s) \equiv
          three\_d\_location(y) \wedge fits(x,y) \wedge exposed(y,s)
\eeq\beq \label{eq:pre-attachB}
     \lefteqn{Poss(attach(x,y),s) \equiv physical\_object(x) \wedge
     physical\_object(y) \wedge} \nl
          & & fits(x,y) \wedge exposed(y,s)
\eeq\beq \label{eq:pre-pourB}
     PossG(pour(x,y),s) \equiv
          raw\_material(x) \wedge three\_d\_location(y) \wedge exposed(y,s)
\eeq\beq \label{eq:pre-removeB}
     Poss(remove(x,y),s) \equiv
          contains(y,x,s) \wedge exposed(x,s)
\eeq\beq \label{eq:pre-pressB}
     Poss(press(x),s) \equiv button(x) \vee lever(x)
\eeq\beq \label{eq:pre-openB}
     Poss(open(x),s) \equiv openable(x)
\eeq\beq \label{eq:pre-closeB}
     Poss(close(x),s) \equiv openable(x)
\eeq\beq \label{eq:pre-touchB}
     Poss(touch(x),s) \equiv
          physical\_object(x) \wedge exposed(x,s)
\eeq\beq \label{eq:pre-getburnedB}
     Poss(get\_burned,s) \equiv
          \exists x,t.(touching(x,s) \wedge temperature(x,t,s) \wedge t \geq
          70)
\eeq\beq \label{eq:pre-steamifyB}
     Poss(\mbox{\it steamify\/}(x),s) \equiv \exists t.(temperature(x,t,s)
     \wedge t \geq 100) \wedge contains(x,water,s)
\eeq

For this domain, it is useful for us to define axiom~(\ref{eq:pre-pourB}) as a
{\it generic\/} action precondition axiom---denoted by {\it PossG}\footnote
     {{\it PossG\/} is defined in the same way as {\it Poss\/} (see
     axiom~(\ref{eq:act-precond-axiom})).}---
because the order in which the ingredients are poured into the baking pan is
important, and the kneading blade should be attached to the baking pan before
any ingredients are poured on.  The {\it specific\/} action precondition axioms
for pouring each ingredient are listed below.\footnote
     {A {\it specialized\/} action is one whose arguments are constant symbols
     rather than variables.}

Axioms~(\ref{eq:pre-openB}) and~(\ref{eq:pre-closeB}) state that an {\it
open\/} or {\it close\/} action on $x$ is possible if $x$ is {\it openable}.
For the modelling of the domain, deciding whether or not something is openable
may be uncertain.  For example, an empty box may or may not be openable
depending on its structure; however, a lid by its very nature is openable,
whereas a sheet of paper is not.

Axiom~(\ref{eq:pre-steamifyB}) expresses that steam is produced in $x$ if $x$
contains water, and the temperature of $x$ is at least 100 \dc.

\paragraph{Domain-specific axioms}

\beq \label{eq:pre-pourwaterB}
     \lefteqn{Poss(pour(water,baking\_pan\_interior),s) \equiv
     PossG(pour(water,baking\_pan\_interior),s) \wedge} \nl
          & & attached(kneading\_blade,baking\_pan,s)
\eeq\beq \label{eq:pre-pourflourB}
     \lefteqn{Poss(pour(flour,baking\_pan\_interior),s) \equiv
     PossG(pour(flour,baking\_pan\_interior),s) \wedge} \nl
          & & contains(baking\_pan,water,s)
\eeq\beq \label{eq:pre-pouryeastB}
     \lefteqn{Poss(pour(yeast,baking\_pan\_interior),s) \equiv
     PossG(pour(yeast,baking\_pan\_interior),s) \wedge} \nl
          & & contains(baking\_pan,flour,s)
\eeq\beq \label{eq:pre-raisetempB}
     \lefteqn{Poss(raise\_temp(x),s) \equiv (x=main\_body \vee x=baking\_pan
     \vee} \nl
          & & vents(x,main\_body) \wedge contains(main\_body,steam,s)) \wedge
          \nl
          & & \exists t.(temperature(x,t,s) \wedge t<200) \wedge
          pressed(on\_button,s)
\eeq\beq \label{eq:pre-flashB}
     \lefteqn{Poss(flash(x),s) \equiv x=complete\_light \wedge} \nl
          & & \exists t.(temperature(baking\_pan,t,s) \wedge t \geq 200)
\eeq

Axiom~(\ref{eq:pre-pourwaterB}) states that the specific action of pouring
water into the baking pan is possible if the generic action is possible, and
the kneading blade is attached to the baking pan.

Axiom~(\ref{eq:pre-pourflourB}) expresses that the specific action of pouring
flour into the baking pan is possible if the generic action is possible, and
the baking pan already contains water.

Thus, axioms~(\ref{eq:pre-pourwaterB}), (\ref{eq:pre-pourflourB}), and
(\ref{eq:pre-pouryeastB}) encode the knowledge that the kneading blade must be
attached to the baking pan before the water, flour, and yeast are poured (in
that order) into the baking pan.

Axiom~(\ref{eq:pre-raisetempB}) asserts that the {\it raise\_temp\/} action can
be performed on the main body, the baking pan, and the steam vent provided that
steam is present inside the main body, that their temperatures are below 200
\dc\ and that the {\small ON} button is pressed.  Raising the temperature of
the steam vent models part of what happens when steam is escaping from the
steam vent---the steam is actually causing much of the temperature change to
occur.  It need not concern us here that we have not included the possibilities
of the {\it raise\_temp\/} action being performed on any other components.

Axiom~(\ref{eq:pre-flashB}) expresses that the ``complete'' light of the
breadmaker flashes when the temperature of the baking pan reaches 200 \dc: this
indicates the end of the baking process, and is, of course, a gross
oversimplification.

\subsubsection{Positive effect axioms}

\paragraph{Domain-independent axioms}

\beq \label{eq:pos-attachedB}
     Poss(a,s) \wedge a=attach(x,y)
          \rightarrow attached(x,y,do(a,s))
\eeq\beq \label{eq:pos-openedB}
     Poss(a,s) \wedge a=open(x)
          \rightarrow opened(x,do(a,s))
\eeq\beq \label{eq:pos-pressedB}
     Poss(a,s) \wedge a=press(x)
          \rightarrow pressed(x,do(a,s))
\eeq\beq \label{eq:pos-touchingB}
     Poss(a,s) \wedge a=touch(x)
          \rightarrow touching(x,do(a,s))
\eeq\beq \label{eq:pos-burnedB}
     Poss(a,s) \wedge a=get\_burned
          \rightarrow burned(do(a,s))
\eeq

\paragraph{Domain-specific axioms}

\beq \label{eq:pos-containsB}
     \lefteqn{Poss(a,s) \wedge (a=insert(x,y) \vee a=pour(x,y) \vee} \nl
          & & a=\mbox{\it steamify\/}(y) \wedge x=steam \vee \nl
          & & a=flash(complete\_light) \wedge x=bread \wedge
          y=baking\_pan\_interior) \rightarrow \nl
          & & contains(y,x,do(a,s))
\eeq\beq \label{eq:pos-exposedB}
     Poss(a,s) \wedge a=open(lid) \wedge contains(main\_body,x,s)
          \rightarrow exposed(x,do(a,s))
\eeq\beq \label{eq:pos-temperature1B}
     \lefteqn{Poss(a,s) \wedge a=raise\_temp(x) \wedge temperature(x,t,s)
     \rightarrow} \nl
          & & temperature(x,t+50,do(a,s))
\eeq\beq \label{eq:pos-temperature2B}
     Poss(a,s) \wedge a=flash(complete\_light)
          \rightarrow temperature(x,20,do(a,s))
\eeq\beq \label{eq:pos-temperature3B}
     Poss(a,s) \wedge a=\mbox{\it steamify\/}(y) \wedge x=steam
          \rightarrow temperature(x,100,do(a,s))
\eeq

Axiom~(\ref{eq:pos-containsB}) expresses that $y$ contains $x$ if $x$ is
inserted or poured into $y$; it also expresses that $y$ contains steam if $y$
gets ``steamified'', and that the interior of the baking pan contains bread
once the baking cycle is complete (indicated by the ``complete'' light
flashing).

We point out here the important difference between the modelling of the
materials in the toaster and breadmaker domains.  For the toaster domain, we
had a single material---the bread slice---for which we used the fluent {\it
toasted\/} to describe its final desired stated (see axiom~\ref{eq:toasted});
however, the bread slice remained a bread slice, and no new object (\ie {\it
toast}) was introduced.  The breadmaker domain is considerably more complex,
however.  There are initially three raw materials in the baking pan---flour,
water, and yeast---and at the end of the baking process, the object {\it
bread\/} is ``created'', and is contained in the baking pan (as described by
axiom~(\ref{eq:pos-containsB}).  Thus, the state changes of the materials are
modelled to some extent.  This approach is necessary for the breadmaker domain,
but it also provides a simple means by which a different linguistic expression
can be used to refer to the final product.  If the state changes of the
materials were similarly modelled in the toaster domain, we would have easily
been able to refer to the final product as {\it the slice of toast\/}, which is
preferable to {\it the bread slice}.\footnote
     {To make this change, axiom~(\ref{eq:pos-contains}) should be modified as
     follows:
     \beq
          \lefteqn{Poss(a,s) \wedge (a=insert(x,y) \vee} \nl
               & & a = raise\_temp(bread\_slice) \wedge
               temperature(bread\_slice,220,do(a,s)) \wedge \nl
               & & \hspace{10mm} x = toast \wedge y = bread\_slot) \rightarrow
               \nl
               & & contains(y,x,do(a,s))
     \eeq}
Note that for a more complex description of the breadmaker domain, an
intermediate state of the materials---dough---may need to be modelled, because
dough may be the final product of the device for some of its program settings.
Also note that according to axiom~(\ref{eq:pos-containsB}), the baking pan
still contains water, flour, and yeast at the end of the baking process.  This
is not important here, but we observe that a decision may sometimes have to be
made as to whether the creation of a composite material should result in the
``destruction'' of its constituent materials.

Axiom~(\ref{eq:pos-exposedB}) states that whatever is contained in the main
body becomes exposed when the lid is opened.

Axiom~(\ref{eq:pos-temperature2B}) asserts that the temperatures of all the
components and materials of the breadmaker system become 20 \dc\ when the bread
is finished.  Axiom~(\ref{eq:pos-temperature3B}) states that the temperature of
the steam, when it is initially produced, is 100 \dc.

\subsubsection{Negative effect axioms}

\paragraph{Domain-independent axioms}

\beq \label{eq:neg-containsB}
     Poss(a,s) \wedge a=remove(x,y)
          \rightarrow \neg contains(y,x,do(a,s))
\eeq\beq \label{eq:neg-attachedB}
     Poss(a,s) \wedge a=remove(x,y)
          \rightarrow \neg attached(x,y,do(a,s))
\eeq\beq \label{eq:neg-openedB}
     Poss(a,s) \wedge a=close(x)
          \rightarrow \neg opened(x,do(a,s))
\eeq

\paragraph{Domain-specific axioms}

\beq \label{eq:neg-pressedB}
     Poss(a,s) \wedge a=flash(complete\_light)
          \rightarrow \neg pressed(on\_button,do(a,s))
\eeq\beq \label{eq:neg-exposedB}
     Poss(a,s) \wedge a=close(lid) \wedge contains(main\_body,x) \rightarrow
     \neg exposed(x,do(a,s))
\eeq

Axiom~(\ref{eq:neg-pressedB}) asserts that the {\small ON} button ceases to be
pressed at the end of baking.

Axiom~(\ref{eq:neg-exposedB}) states that whatever is contained in the main
body of the breadmaker is not exposed any more after the lid is closed.

\subsection{A sample generated instruction sequence}

If we define the bread to be finished when the ``complete'' light starts
flashing, we can set the goal G of the planner to be this:
\be
     G = finished(bread) \wedge removed(bread,baking\_pan\_interior)
\ee

The final sequence of instructions, generated by Penman, including warning
instructions, is the following (see appendix~\ref{app:a} for the full output of
the system):
\begin{small}
\begin{verbatim}
     Attach the kneading blade to the baking pan.
     Pour the water into the baking pan.
     Pour the flour into the baking pan.
     Pour the yeast into the baking pan.
     Insert the baking pan into the main body.
     Close the lid.
     Press the ON button.
     Do not touch the main body during the heating period.
     Do not touch the steam vent during the heating period.
     The ``complete'' light will flash.
     Open the lid.
     Take the baking pan out of the main body.
     Take the bread out of the baking pan.
\end{verbatim}
\end{small}

\subsection{Combining the breadmaker and toaster domains}

By allowing our system access to the axioms for both the breadmaker and toaster
domains, and adding the {\it slice}\footnote
     {This action is lexicalized as ``cut'' in our system.}
action that can be performed by the user on the bread to produce a bread slice,
the system can generate a sequence of sentences which instruct the user how to
obtain a slice of toast starting from the ingredients for bread (see
appendix~\ref{app:a}):
\begin{small}
\begin{verbatim}
     Attach the kneading blade to the baking pan.
     Pour the water into the baking pan.
     Pour the flour into the baking pan.
     Pour the yeast into the baking pan.
     Insert the baking pan into the main body.
     Close the lid.
     Press the ON button.
     Do not touch the main body during the heating period.
     Do not touch the steam vent during the heating period.
     The ``complete'' light will flash.
     Open the lid.
     Take the baking pan out of the main body.
     Take the bread out of the baking pan.
     Cut the bread slice from the bread.
     Insert the bread slice into the toaster's bread slot.
     Press the ON lever.
     Do not touch the toaster's bread slot during the heating period.
     The bread slice will pop up.
     Take the bread slice out of the toaster's bread slot.
\end{verbatim}
\end{small}

\chapter{Discussion and conclusions}

\section{An integrated approach to device design and instruction generation}

In chapter~3 we argued that a complete natural language instruction generation
system should incorporate topological, kinematic, electrical, thermodynamic,
electronic, and world knowledge at the top level, on which all the necessary
reasoning to arrive at the final instructions is carried out.  We propose here
a general framework within which a device may be designed by the engineer,
with one of the intended ``side effects'' being the generation of instructions
to perform a given task.  This methodology starts with an engineering approach
to the design of a kitchen appliance.  The possible uses of the design
components with respect to natural language generation will be considered.

\subsection{The design phase}

In this section, we provide an overview of the steps that we propose could be
taken as part of designing a kitchen appliance:
\begin{enumerate}
     \item For the device, construct solid, kinematic, electrical,
     thermodynamic, and electronic models.
     \item Determine the {\it salient states} of the components of the device.
\end{enumerate}

\subsubsection{Construction of the device models}

There are industrial packages which allow an engineer to construct solid and
kinematic models of a device.  A solid model defines the topology of the
device, whereas a kinematic model describes the motions of movable components.
Electrical modelling software also exists; it should allow one to formally
describe flow of charge and resistance.  We are not aware of any thermodynamic
modelling software currently in existence, but if and when this comes into
existence, it should enable the engineer to specify the {\it materials\/}
comprising each component and connection.  In conjunction with the electrical
model, the thermodynamic model should allow the temperatures of various
components to be estimated at any given time.

\subsubsection{Identification of component states}

During the construction of the models mentioned above, the {\it salient
states\/} of each component will have been identified.  For example, although
the kinematics component of the integrated model may describe a {\it lid\/} as
having a certain range of motion, it is probably not necessary to regard every
position of the lid as a separate state, because the difference between one
position and another neighbouring one may not have any effect on the rest of
the device--environment system.  However, when the lid gets to a certain
position, let's call this {\it closed}, another component of the system may
become enabled.  Thus, we consider the {\it closed\/} position of the lid to
be a {\it salient state\/} of the lid.  We envision that the enabling of one
component by the closing of the lid should be part of the integrated model.\\

Note that these stages are concerned with the {\it device\/} rather than the
{\it device--environment system}.  This is because: (1) the engineer is
presumably building a model of the device and not the environment, and (2) we
believe that the environment model can be very general to a large extent, \ie
domain-independent.  The environment model should model possible user
interactions with any device, such as the pushing of buttons, the touching of
components, the cutting of materials, etc.

\subsection{Incorporating instruction generation into the framework}

The design phase can be succeeded by the following steps leading to the
generation of natural language instructions:
\begin{enumerate} \setcounter{enumi}{2}
     \item Determine the actions and fluents for this domain.
     \item Derive the action precondition axioms and the positive and negative
     effect axioms for this domain, using the design specification and world
     knowledge.
     \item \label{en2:1} Determine the goal of the system, and plan a sequence
     of actions leading to the goal becoming true.
     \item Perform further in-line planning, using the basic plan, to
     determine potentially dangerous states.
     \item Determine the relevant case roles for the actions.
     \item Decide which actions should be mentioned.
     \item \label{en2:2} Generate a PRL (Process Representation Language)
     specification for these actions.
     \item Determine the features of the system from the design specification
     and world knowledge, and use these as answers to IMAGENE inquiries.
     \item Feed the PRL specification and the results of the inquiries into
     IMAGENE.
\end{enumerate}
Steps~(\ref{en2:1})--(\ref{en2:2}) are what this thesis concentrates on.
However, instead of generating a PRL specification for the text, a SPL
(Sentence Plan Language) expression is produced, which is fed into Penman (see
section~\ref{sec:generate}).

\subsubsection{Determination of domain-specific actions and fluents}

The domain fluents are simply symbols assigned to the salient states, which
were identified in step~(2).

In section~\ref{sec:ontology}, we outlined a basic ontology of the high-level
device actions to be represented.  The particular device actions used depend
on the kinematic, electrical, and thermodynamic models---for instance, a
washing machine needs an action symbol representing {\it rotate}, while a
toaster will not.  Also, the number and meanings of the arguments of these
actions must be determined.  As we saw in section~\ref{sec:determineroles},
these arguments specify some of the roles of the action.  However, some device
actions, such as {\it pop\_up}, have no arguments, and thus a role may need to
be inferred from the plan if the action is to be mentioned.  A special
domain-specific program clause needs to be added for this purpose.  For
example, to determine what pops up from the bread slot at a given point in the
plan, the clause must find out what the bread slot contains at that point.

As we have already mentioned, we believe that the user interactions with a
device are largely domain-independent; thus, the reader actions are
transferable to other domains.  Although there are some domain-dependent {\it
axioms}, such as the specific action precondition axioms for the {\it pour\/}
action (defined in section~\ref{sec:breadmaker-axioms}), the symbols
representing these actions should still be domain-independent.

\subsubsection{Derivation of domain-specific axioms from design and world
knowledge}
\label{sec:deriveaxioms}

In this section we will not propose a general mechanism by which the
domain-specific axioms can be derived from the device and environment models,
but instead we will try to justify the derivability of some of them from these
models.  We will be referring to axioms presented in
sections~\ref{sec:toaster-axioms} and \ref{sec:breadmaker-axioms}, and to the
knowledge types outlined in section~\ref{sec:knowledgetypes}).

\paragraph{Axioms~(\ref{eq:pre-raisetemp}) and~(\ref{eq:pos-temperature1})}
There is more than one possible way in which axiom~(\ref{eq:pre-raisetemp})
can hold; we shall consider one for now.  We can assume that the correctness
of the axiom is dictated by the electrical subsystem (\ie the electrical
model), the thermodynamic properties of the components and materials of the
system (\ie the thermodynamic model), and the topology of the system
(\ie the solid model).  When the {\small ON} button is pressed, a switch
allows an electrical current to begin flowing through the heating element.
The resistance of the electrical components together with the current will
determine the maximum temperature reachable by any component.  Also, the
thermodynamic properties of the components connected to, or near to, the
heating element determine the maximum temperature reachable by those
components.  For example, when a slice of bread is in the bread slot, it gets
heated mainly by radiation of heat from the heating element, and to a much
lesser extent, conduction of heat through the air between the bread and the
heating element.  These physical processes can be represented, to some degree
of accuracy, by equations.

Axiom~(\ref{eq:pre-raisetemp}) can be elaborated by examining the solid and
thermodynamic models, and considering which components other than the bread
slot and its contents can become heated by the heating element.  The {\it
raise\_temp\/} action may be applicable to other parts as well, but we have
only considered one component and one material for simplicity.

Axiom~(\ref{eq:pos-temperature1}) is an approximation of a physical equation,
and a conversion must be made from the continuous equation into the discrete
axiom.

\paragraph{Axiom~(\ref{eq:pre-popup})}
This axiom is justifiable if the toaster contains a thermostat which senses
the temperature of the bread slot.  The solid model will specify the location
of the thermostat relative to the other components, and the electrical and
thermodynamic models will specify its functionality.  That is, when the
thermostat senses a temperature of 200 \dc\ in the bread slot, it triggers the
popping up action.

\paragraph{The {\it pop\_up\/} action}
This action causes a number of things to happen (\ie changes the truth values
of fluents).  Whatever was in the bread slot becomes exposed; this can be
determined from the solid and kinematic models.  This action causes the state
of the {\small ON} button to become not pressed, \ie it ``breaks the
connection'' in the electrical circuit between the {\small ON} button and the
electrical subsystem.  Also, it marks the beginning of the cooling down period
of every component as a result of the switching off of the electrical current.
We could have modelled this continuous cooling down period with more axioms
representing discrete temperature changes, but to make matters simple, we
ignored this period altogether and assumed that the cooling down is
instantaneous.

\paragraph{Axioms~(\ref{eq:pre-pourwaterB}), (\ref{eq:pre-pourflourB}), and
(\ref{eq:pre-pouryeastB})}
The {\it pour\/} action is an example of an action that is domain-independent,
but whose {\it specializations\/} are not, in the case of the breadmaker.  The
fact that the baking pan must contain water before the flour is poured, and
that the flour must be present before the yeast is poured, should be part of
world knowledge: there should be a fact stating that the yeast must not come
directly into contact with the water, or else the yeast may not perform its
function (in the rising of the dough) properly.\footnote
     {Why the yeast, flour, and water should not be poured in that order is
     another point to analyze.  Presumably, the kneading blade cannot mix the
     ingredients well if it is initially surrounded by the dry ingredients.}

One may wonder why axiom~(\ref{eq:pre-pourwaterB}) needs to specify that the
kneading blade be attached to the baking pan before the water can be poured;
deriving this axiom {\it could\/} use some world knowledge about it being
difficult to attach the kneading blade once there are ingredients already in
the baking pan.  A more realistic approach would have been to define the {\it
bottom of the baking pan\/} as a physical object and asserting
$fits(kneading\_blade,baking\_pan\_bottom)$ rather than
$fits(kneading\_blade,baking\_pan)$, so that when an ingredient is poured into
the baking pan, the {\it baking\_pan\_bottom\/} is no longer exposed; thus the
$kneading\_blade$ cannot now be attached to the $baking\_pan\_bottom$.
Although the way this is currently implemented is simpler than the more
realistic approach, this example serves to illustrate that it will sometimes
be difficult to decide whether {\it parts\/} of the physical objects of the
solid model should be used in constructing the axioms.

\subsubsection{The procedural planner}

The planner employed in our current implementation is a very basic
forward-chaining planner that attempts to reach the goal state from the
initial state.  It would have been preferable to have implemented a regression
planner such as that of Lin \shortcite{lin95}; all of the axioms presented
in chapter~\ref{ch:main} would still be correct, but their forms would have to
be modified for use with the given planner.  However, this thesis is not
concerned with any particular planning paradigm.  We just need to make the
reader aware that several of the action precondition axioms used in the
implementation have extra conditions which were added only to allow the
current planner to function correctly.  A linear regression planner would not
need these extra conditions.

\subsubsection{Determining what actions to mention}

In section~\ref{sec:otheractions} we looked at some situations in which
certain actions should or should not be mentioned.  We propose that a study
essentially similar to that of Vander Linden's \shortcite{vanderlinden93b}
should be undertaken to determine how the {\it features\/} of the environment
and communicative context affect the inclusion of actions in instructional
text.

\subsubsection{Integration with IMAGENE}

Vander Linden's PRL is rather similar to the basic form of Penman's SPL that
our system produces, in that many of the case roles determined by our system
are also used by PRL (see sections~\ref{sec:vanderlinden}
and~\ref{sec:determineroles})

In order to take full advantage of IMAGENE's expressiveness (because several
of the features of instructional text identified by Vander Linden are based on
action hierarchies and concurrency), a hierarchical planner should be
implemented.  Vander Linden acknowledges that as well as the planner needing
to determine the content of the instructional text, it would ``also be
critical in implementing the text-level inquiries, a step that is required for
fully automating the instruction generation process'' \cite[page
133]{vanderlinden93b}.  Thus the planner, or a system working in tandem with
the planner\footnote
     {Such a planner might possibly be akin to Kosseim and Lapalme's {\it
     semantic level\/} (see section~\ref{sec:kosseim&lapalme}).},
should be able to identify these features (the values of which are input to
IMAGENE via its inquiries).

We propose that several of these features should be identifiable from the
knowledge in the solid and kinematic models, and some are already related to
the fluents we have used in our toaster and breadmaker examples:
\begin{description}
     \item [Action-Actor] Is the action being performed by the reader or some
     other agent?
     \item [Action-Monitor-Type] Is this non-reader action expected to be
     monitored by the reader?
     \item [Precond-Inception-Status] Could the reader have witnessed the
     inception of the process on which the precondition is being based?
\end{description}
Whether an action is expected to be monitored or witnessed by the reader is
related to the use of our {\it exposed\/} fluent, and the idea of a {\it
salient change\/} used in our implementation (see
section~\ref{sec:otheractions}).

\subsubsection{Representation of continuous time and physical processes}

The situation calculus formalism we have used does not have any explicit
representation of time.  If we are to fully capture the temporal relationships
between actions and address the time-related features of instructional text
identified by Vander Linden, we will need to use a formalism that allows the
representation of time explicitly, such as that of Pinto \shortcite{pinto94}.
For instance, PRL allows the specification of the role {\small DURATION} for
an action (see section~\ref{sec:vanderlinden}), and the features queried about
by IMAGENE include the following:
\begin{description}
     \item [Temporal-Orientation] Is the action one which was performed at a
     temporally remote time in the past?
     \item [Concurrency-Structure] Is the action a procedure with concurrency
     that must be expressed?
\end{description}

We suspect that using a formalism that allows continuous equations (to
represent the physical processes), such as those of Sandewall
\shortcite{sandewall89} and Levesque and Reiter \shortcite{levesque&reiter95},
would make it more straightforward for the axioms to be derived, because the
meanings of the axioms would then be ``closer'' to the device model.  However,
reasoning in those formalisms is much more complex than in the basic situation
calculus.

\section{Contributions of this thesis}

This thesis showed how it is possible to go from a model of a kitchen
appliance, characterized by axioms in the situation calculus, to the
generation of natural language instructions which explain the steps the user
should take to operate the device as well as instructions which warn the user
to avoid potentially dangerous situations.  The behaviour of the appliance is
simulated by the planning mechanism, which attempts to determine all the
situations that are potentially hazardous to the user.

The contributions of this thesis are therefore:
\begin{enumerate}
     \item the suggestion that an integrated model of the device (including
     solid, kinematic, electrical, and thermodynamic models) together with
     world knowledge can be used to automate the generation of instructions,
     including warning instructions;
     \item that situations in which injuries to the user can occur need to be
     planned for at every step in the planning of the {\it normal\/} operation
     of the device, and that these ``injury sub-plans'' are used to instruct
     the user to avoid these situations.  Thus, unlike other instruction
     generation systems, our system tells the reader what {\it not\/} to do as
     well as what to do; and
     \item the notion that actions are performed on the materials that the
     device operates upon, that the states of these materials may change as a
     result of these actions, and that the goal of the system should be
     defined in terms of the final states of the materials.
\end{enumerate}

\appendix
\chapter{Program listing}

This appendix contains a listing of the Quintus Prolog program which
implements the ideas presented in chapter 4, for the toaster domain.

For the domain model, clauses which have extra conditions needed only for
the current planner are marked with a \verb"/*!*/" on the right side of
the page.\\

\begin{footnotesize}
\begin{verbatim}
:- no_style_check(all).

/* DOMAIN DESCRIPTION */

/* Preconditions for actions */

poss(insert(X,Y),S) :-
   fits(X,Y), three_d_location(Y),
   holds(exposed(Y),S),
   \+ holds(contains(Y,X),S).                                          /*!*/

poss(remove(X,Y),S) :-
   three_d_location(Y),
   holds(contains(Y,X),S),
   holds(exposed(X),S).

poss(press(X),S) :-
   lever(X),
   \+ holds(pressed(X),S).                                             /*!*/

poss(raise_temp(X),S) :-
   (X=bread_slot; holds(contains(bread_slot,X),S)),
   holds(temperature(X,T),S), T < 200,
   holds(pressed(on_lever),S).

poss(pop_up,S) :-
   holds(temperature(bread_slot,T),S), T >= 200.

poss(get_burned,S) :-
   holds(touching(X),S),
   holds(temperature(X,T),S), T >= 70.

poss(touch(X),S) :-
   physical_object(X), holds(exposed(X),S),
   holds(temperature(X,T),S), T > 20.                                  /*!*/


/* Successor state axioms */

holds(contains(Y,X),do(A,S)) :-
   A = insert(X,Y);
   \+ A = remove(X,Y), holds(contains(Y,X),S).

holds(removed(X,Y),do(A,S)) :-
   A = remove(X,Y);
   holds(removed(X,Y),S).

holds(pressed(X),do(A,S)) :-
   A = press(X);
   \+ A = pop_up, holds(pressed(X),S).

holds(exposed(X),do(A,S)) :-
   X = bread_slot;                                                     /*
Always exposed */
   A = pop_up, holds(contains(bread_slot,X),S);
   \+ A = press(on_lever), holds(exposed(X),S).

holds(temperature(X,T2),do(A,S)) :-
   A = raise_temp(X), holds(temperature(X,T1),S), T2 is T1+50;
   A = pop_up, T2 is 20;
   \+ A = raise_temp(X), \+ A = pop_up, holds(temperature(X,T2),S).

holds(burned,do(A,S)) :-
   A = get_burned;
   holds(burned,S).

holds(touching(X),do(A,S)) :-
   A = touch(X);
   holds(touching(X),S).

holds(toasted(X),do(A,S)) :-
   holds(temperature(X,220),do(A,S));
   holds(toasted(X),S).


/* Initial state */

holds(temperature(bread_slice,20),s0).
holds(temperature(bread_slot,20),s0).
holds(exposed(bread_slot),s0).
holds(exposed(bread_slice),s0).


/* General */

physical_object(bread_slot).
physical_object(on_lever).
three_d_location(bread_slot).
fits(bread_slice,bread_slot).
lever(on_lever).
raw_material(bread_slice).
indicator(nothing).                                                    /* Not
used for this domain */

reader_action(insert).
reader_action(remove).
reader_action(press).
reader_action(touch).
device_action(raise_temp).
device_action(pop_up).

actor(flash(X),X).

actee(insert(X,_),X).
actee(remove(X,_),X).
actee(press(X),X).
actee(touch(X),X).
actee(raise_temp(X),X).

source(remove(_,Y),Y).

destination(insert(_,Y),Y).

polarity(touch(_),P) :-
   !, P = negative.
polarity(A,positive).

normal_action(A) :-                                                    /*!*/
   A =.. [Action|Args],                                                /*!*/
   member(Action,[insert,remove,press,remove,raise_temp,pop_up]).      /*!*/

injury_action(A) :-                                                    /*!*/
   A =.. [Action|Args],                                                /*!*/
   member(Action,[touch,get_burned]).                                  /*!*/

affects(insert(X,Y), contains(Y,X)).                                   /*!*/
affects(remove(X,Y), removed(X,Y)).                                    /*!*/
affects(remove(X,Y), contains(Y,X)).                                   /*!*/
affects(press(X), pressed(X)).                                         /*!*/
affects(press(on_lever), exposed(X)).                                  /*!*/
affects(get_burned, burned).                                           /*!*/
affects(touch(X), touching(X)).                                        /*!*/


/* DOMAIN-INDEPENDENT CLAUSES */

/* Main clause */

run :-
   planNormal(s0,G,[toasted(bread_slice), removed(bread_slice,X)]),
   listActions(G,L1),
   indexActions(L1,1,L2),
   writeln('Inserting injuries...'),
   insertInjuries(L2,L3),
   write('WITH INJURIES: '), write(L3), nl,
   writeln('Making interpretations...'),
   makeInterpretations(L3,L4,Patterns),
   write('INTERPRETATIONS: '), write(L4), nl,
   write('PATTERNS: '), write(Patterns), nl,
   writeln('Making SPL...'),
   makeSPL(L4,L4,Patterns,SPL,0),
   outputSPL0(SPL),
   writeln('Done.').


/* The forward planner */

planNormal(Goal_state,Goal_state,Goals) :-
   satisfied(Goals,Goal_state),
   write('GOAL STATE: '), write(Goal_state), nl.

planNormal(Current_state,Goal_state,G) :-
   poss(A,Current_state),
   normal_action(A),
   \+ loop(Current_state,A),                                           /* Try
to avoid infinite loop */
   planNormal(do(A,Current_state),Goal_state,G).

planInjury(Goal_state,Goal_state,Goals) :-
   satisfied(Goals,Goal_state).

planInjury(Current_state,Goal_state,G) :-
   poss(A,Current_state),
   injury_action(A),
   \+ loop(Current_state,A),
   planInjury(do(A,Current_state),Goal_state,G).

satisfied([],Goal_state) :- !.

satisfied([G|Goals],Goal_state) :-
   holds(G,Goal_state),
   satisfied(Goals,Goal_state).

loop(do(A1,S),A2) :-
   affects(A1,F), affects(A2,F).

/* Find all points in the plan which can lead to an injury */

insertInjuries(L1,L2) :-
   write('INDEX: '),
   getInjuryPoints(L1,1,L3),
   write('POINTS: '), write(L3), nl,
   mergeActions(L1,L3,L2).

getInjuryPoints(L,I,[]) :-
   \+ member((I,_),L),
   nl, !.

getInjuryPoints( L1, I, [(I,L8)|L2]) :-
   getFirst(L1,I,L3),                                                  /* Get
first I indexed actions */
   deListify(L3,L4),
   makeState(L4,S),
   write('['), write(I), write('] '),
   planInjury(S,G,[burned]),                                           /*
Invoke planner */
   listActions(G,L5),                                                  /*
Listify goal state */
   indexActions(L5,1,L6),
   getLast(L6,I,L7),                                                   /* Keep
actions after I */
   collectActions(L7,L8),                                              /*
Remove indices */
   J is I+1,
   getInjuryPoints(L1,J,L2).

getInjuryPoints(L1,I,L2) :-
   J is I+1,
   getInjuryPoints(L1,J,L2).

mergeActions(L,[],L).

mergeActions( [(I1,[A])|L1], [(I1,L2)|L3], [(I1,[A|L2])|L4]) :-
   mergeActions(L1,L3,L4).

mergeActions([P|L1],L2,[P|L3]) :-
   mergeActions(L1,L2,L3).

/* Make interpretations */

makeInterpretations(L1,L2,Patterns) :-
   makeInts(L1,L3,Patterns),
   splitGroup(L3,L4),
   deListify(L4,L2).

makeInts(L1,L2,[Pattern|Patterns]) :-
   getFirstOccurrence(L1,raise_temp,I),                                /* Find
first action in */
   getTail(L1,I,L3),                                                   /*
heating pattern */
   checkPattern(L3,raise_temp,I,J,Pattern),                            /* Get
rest of actions */
   reIndexSame(L3,I,J,InGrp),                                          /* Make
actions in group have */
                                                                       /*  same
index */
   removeSuperfluousActions(InGrp,[],InGroup),                         /*
Remove duplicate injury actions */
   write('InGroup: '), write(InGroup), nl,
   getLast(L1,J,L4),
   H is I+1,
   reIndexIncrementing(L4,H,AfterGroup),                               /*
Reindex actions after group */
   K is I-1,                                                           /*  in
ascending order */
   getFirst(L1,K,BeforeGroup),
   append(BeforeGroup,InGroup,BeforeAndInGroup),
   makeInts(AfterGroup,NewAfterGroup,Patterns),
   append(BeforeAndInGroup,NewAfterGroup,L2).

makeInts(L,L,[]).

getFirstOccurrence(L,Act,I) :-
   appendz( L1, [ (I,[A|Actions]) | IActions ], L),
   A =.. [Act|Args].

checkPattern( [(I,[A|Actions])|IActions], Act, J, K, Pattern) :-
   A =.. [Act|Args],
   checkPattern(IActions,Act,J,K,Pattern).

checkPattern( [(I,[A|Actions])], Act, J, I, (J,heating_period)) :-
   A =.. [Act|Args],
   I > J+2.

checkPattern( [(I,Actions)|IActions], Act, J, K, (J,heating_period)) :-
   I > J+2,                                                            /*
Assign label only if length */
   K is I-1.                                                           /*  of
collection is at least 3 */

reIndexSame( [(I,Actions)|IActions], J, I, [(J,Actions)]) :- !.

reIndexSame( [(I,[A|Actions])|IActions1], J, K,
  [(J,[A|Actions])|IActions2]) :-
   reIndexSame(IActions1,J,K,IActions2).

reIndexIncrementing([],I,[]) :- !.

reIndexIncrementing( [(I,[A|Actions])|IActions1], J,
  [(J,[A|Actions])|IActions2]) :-
   M is J+1,
   reIndexIncrementing(IActions1,M,IActions2).

indexSame([],_,[]).

indexSame([A|Actions],I,[(I,[A])|Rest]) :-
   indexSame(Actions,I,Rest).

splitGroup([],[]).

splitGroup([(I,Actions)|IActions],L) :-
   indexSame(Actions,I,L1),
   splitGroup(IActions,L2),
   append(L1,L2,L).

removeInjuries(Actions,InjuryList,InjuryList,[]) :-
   member(Actions,InjuryList),
   !.

removeInjuries(Actions,InjuryList,[Actions|InjuryList],Actions).

removeSuperfluousActions( [(I,[A])|IActions], InjuryList, [(I,[A])|Rest]) :-
   removeSuperfluousActions(IActions,InjuryList,Rest).

removeSuperfluousActions( [(I,[A|Actions1])|IActions], InjuryList,
  [(I,[A|Actions2])|Rest]) :-
   removeInjuries(Actions1,InjuryList,NewInjuryList,Actions2),
   removeSuperfluousActions(IActions,NewInjuryList,Rest).

removeSuperfluousActions([],_,[]).


/* Make SPL */

makeSPL( [ (I1,A1),(I2,A2),(I3,A3)|IActions], AllActions, Patterns,
  [(ID1,Act2,SF_Pairs1), (ID3,Act3,SF_Pairs2) |L], ID) :-
   nonvar(Patterns),
   I2 is I1+1,
   A1 =.. [Act1|Args1],
   device_action(Act1),
   A2 =.. [Act2|Args2],
   device_action(Act2),
   A3 =.. [Act3|Args3],
   reader_action(Act3),
   caused_salient_change(I2,AllActions),                               /* A2
caused a salient change */
   member((I1,Interpretation),Patterns),                               /* A1 is
last continuous */
   ID1 is ID+1,                                                        /*
action in a collection */
   getSFPairs((I2,A2),AllActions,Patterns,SF_Pairs1,ID1,ID2),          /* Get
roles of action A2 */
   ID3 is ID2+1,
   getSFPairs((I3,A3),AllActions,Patterns,SF_Pairs2,ID3,ID4),          /* Get
roles of action A3 */
   makeSPL(IActions,AllActions,Patterns,L,ID4).

makeSPL([(I,A)|IActions],AllActions,Patterns,[(ID1,Act,SF_Pairs)|L],ID) :-
   A =.. [Act|Args],
   reader_action(Act),                                                 /*
Normally just mention */
   ID1 is ID+1,                                                        /*
reader actions */
   getSFPairs((I,A),AllActions,Patterns,SF_Pairs,ID1,ID2),
   makeSPL(IActions,AllActions,Patterns,L,ID2).

makeSPL([(_,_)|IActions],AllActions,Patterns,L,ID) :-
   makeSPL(IActions,AllActions,Patterns,L,ID).

makeSPL([],_,_,[],_).

caused_salient_change(I,Actions) :-
   getFirst(Actions,I,L),
   makeState(L,do(A,S)),
   changed_salient(A,S).

changed_salient(A,S) :-
   (physical_object(X); raw_material(X)),
   \+ holds(exposed(X),S),
   holds(exposed(X),do(A,S)).

getSFPairs((I,A),AllActions,Patterns,SF_Pairs,ID1,ID2) :-
   getActor(I,A,AllActions,SF1,ID1,ID3),
   getActee(A,SF2,ID3,ID4),
   getSource(A,SF3,ID4,ID5),
   getDestination(A,SF4,ID5,ID6),
   getTime(I,A,Patterns,SF5,ID6,ID7),
   getTense(A,SF6,ID7,ID8),
   getSpeechact(A,SF7,ID8,ID2),                                        /*
Remove roles if they */
   removeNone([SF1,SF2,SF3,SF4,SF5,SF6,SF7],SF_Pairs).                 /*  have
no filler */

/* Domain-specific clause: determines what pops up by examining what is
   contained in the bread_slot at that point */

getActor(I, pop_up, Actions,
  (actor,(ID2,Actor,[(determiner,(ID2,the,[]))])),
  ID1, ID2) :-
   getFirst(Actions,I,L),
   makeState(L,S),
   holds(contains(bread_slot,Actor),S),
   ID2 is ID1+1.

getActor(_, A, _,
  (actor,(ID2,Actor,[(determiner,(ID2,the,[]))])),
  ID1, ID2) :-
   actor(A,Actor),
   ID2 is ID1+1.

getActor(_, A, _,
  (actor,(hearer,person,[])),
  ID, ID).

getActee(A,
  (actee,(ID2,Actee,[(determiner,(ID2,the,[]))])),
  ID1, ID2) :-
   actee(A,Actee),
   ID2 is ID1+1.

getActee(_,none,ID,ID).

getSource(A,
  (source,(ID2,Source,[(determiner,(ID2,the,[]))])),
  ID1, ID2) :-
   source(A,Source),
   ID2 is ID1+1.

getSource(_,none,ID,ID).

getDestination(A,
  (destination,(ID2,Destination,[(determiner,(ID2,the,[]))])),
  ID1, ID2) :-
   destination(A,Destination),
   ID2 is ID1+1.

getDestination(_,none,ID,ID).

getTime(I,A,Patterns,
  (exhaustive-duration,(ID2,Time,[(determiner,(ID2,the,[]))])),
  ID1, ID2) :-
   theTime(I,A,Patterns,Time),
   ID2 is ID1+1.

getTime(_,_,_,none,ID,ID).

/* We make a simplification here that the tense of a device action is
   always future, because the only time it is mentioned is when it is
   the last action in a collection */

getTense(A,
  (tense,(ID,future,[])),
  ID, ID) :-
   A =.. [Action|Args],
   device_action(A).

getTense(A,
  (tense,(ID,present,[])),
  ID, ID).

/* We also make a simplification that the speechact of a device action
   is always an assertion, for the same reason */

getSpeechact(A,
  (speechact,(ID,assertion,[])),
  ID, ID) :-
   A =.. [Action|Args],
   device_action(A).

getSpeechact(A,
  (speechact,(ID,imperative,[])),
  ID, ID) :-
   polarity(A,positive).

getSpeechact(A,
  (speechact,(ID,neg-imperative,[])),
  ID, ID) :-
   polarity(A,negative).

/* Set the time of an action if it is part of a collection */

theTime(I,A,Patterns,Time) :-
   nonvar(Patterns),
   member((I,Time),Patterns).

removeNone([none|SF_Pairs1],SF_Pairs2) :-
   removeNone(SF_Pairs1,SF_Pairs2).

removeNone([SF|SF_Pairs1],[SF|SF_Pairs2]) :-
   removeNone(SF_Pairs1,SF_Pairs2).

removeNone([],[]).


/* Output SPL */

outputSPL0( SPL ) :-
   tell('toast.spl'),
   write( '(setq plan '), put(39), write( '(' ),
   outputSPL( SPL ),
   write( '))' ), nl,
   told.

outputSPL( [Sentence | Rest] ) :-
   outputSentence( Sentence, 0 ),
   outputSPL( Rest ).

outputSPL( [] ).

outputSentence( (ID, Top_level, SF_pairs), T ) :-
   tab(T), write( '(ID' ),
   write( ID ),
   write( ' / ' ),
   write( Top_level ),
   NewT is T+8,
   outputSF( SF_pairs, NewT ),
   write( ')' ), nl.

outputSF( [(actor,(hearer,person,[])) | Rest], T ) :-
   nl, tab(T), write( ':' ),
   write( 'actor (hearer / person)' ),
   outputSF( Rest, T ).

outputSF( [(Slot,(ID,Filler,[])) | Rest], T ) :-
   nl, tab(T), write( ':' ),
   write( Slot ),
   write( ' ' ),
   write( Filler ),
   outputSF( Rest, T ).

outputSF( [(Slot,(ID,Filler,SF_pairs)) | Rest], T ) :-
   nl, tab(T), write( ':' ),
   write( Slot ),
   write( ' (ID' ),
   write( ID ),
   write( ' / ' ),
   write( Filler ),
   NewT is T+8,
   outputSF( SF_pairs, NewT ),
   write( ')' ),
   outputSF( Rest, T ).

outputSF( [], _ ).


/* Generic clauses */

listActions(do(A,s0),[A]).

listActions(do(A,S),L) :-
   listActions(S,L1),
   append(L1,[A],L).

makeState([(I,A)],do(A,s0)).

makeState(L,do(A,S)) :-
   append(L1,[(I,A)],L),
   makeState(L1,S).

indexActions([],_,[]).

indexActions([A|Actions],I,[(I,[A])|Rest]) :-
   J is I+1,
   indexActions(Actions,J,Rest).

getFirst(L1,I,L2) :-
   append(Before,[(I,Actions)|After],L1),
   append(Before,[(I,Actions)],L2).

getLast(L1,I,L2) :-
   append(Before,[(I,Actions)|L2],L1).

getTail(L1,I,[(I,Actions)|L2]) :-
   append(Before,[(I,Actions)|L2],L1).

collectActions([],[]).

collectActions( [(_,[A])|Rest], [A|L]) :-
   collectActions(Rest,L).

deListify([],[]).

deListify( [(I,[A])|IActions], [(I,A)|Rest]) :-
   deListify(IActions,Rest).

deListify( [(I,[A|Actions])|IActions], [(I,A)|Rest]) :-
   deListify( [(I,Actions)|IActions], Rest).

member( X, [X|_] ).
member( X, [_|L] ) :- member( X, L ).

appendz( [], X, X ).
appendz( [X|Z], Y, [X|Z1] ) :- appendz( Z, Y, Z1 ).

writeln( X ) :- write( X ), nl.
\end{verbatim}
\end{footnotesize}

\chapter{Trace output}
\label{app:a}

This appendix contains a trace of the run for each of the toaster,
breadmaker, and toaster/breadmaker combination domains.

The Prolog program is first invoked; this outputs the SPL for the
instructions to a file, which is read from Penman.

The \verb"GOAL STATE" is the final state produced by the
planner; \verb"POINTS" specifies all the places in the
plan which can lead to an injury; \verb"WITH INJURIES"
shows all the actions with these injury sub-plans included;
\verb"INTERPRETATIONS" shows the actions after the
interpretations have been made and the superfluous injury sub-plans
removed; and the \verb"PATTERNS" represent the periods
of continuous actions inferred during the interpretation stage.

\verb"comb.loom" is the Penman domain model;
\verb"lexicon" specifies the lexical features of each
word linked to the domain model; and \verb"toast.spl",
\verb"bread.spl", and \verb"comb.spl" are
the SPL files.

\section{Output for the toaster domain}
\vspace{4mm}
\begin{footnotesize}
\begin{verbatim}
spawn prolog
Quintus Prolog Release 3.2 (Sun 4, SunOS 5.3)
Copyright (C) 1994, Quintus Corporation.  All rights reserved.
301 East Evelyn Ave, Mountain View, California U.S.A. (415) 254-2800
Licensed to Dept. of Computer Science, University of Toronto

| ?- ['~/prolog/toast'].
% compiling file /homes/neat/a/da/prolog/toast.pl
% toast.pl compiled in module user, 1.240 sec 16,420 bytes

yes
| ?- run.
GOAL STATE: do(remove(bread_slice,bread_slot),do(po
p_up,do(raise_temp(bread_slice),do(raise_temp(bread
_slice),do(raise_temp(bread_slice),do(raise_temp(br
ead_slice),do(raise_temp(bread_slot),do(raise_temp(
bread_slot),do(raise_temp(bread_slot),do(raise_temp
(bread_slot),do(press(on_lever),do(insert(bread_sli
ce,bread_slot),s0))))))))))))
Inserting injuries...
INDEX: [1] [2] [3] [4] [5] [6] [7] [8] [9] [10] [11
] [12]
POINTS: [(3,[touch(bread_slot),get_burned]),(4,[tou
ch(bread_slot),get_burned]),(5,[touch(bread_slot),g
et_burned]),(6,[touch(bread_slot),get_burned]),(7,[
touch(bread_slot),get_burned]),(8,[touch(bread_slot
),get_burned]),(9,[touch(bread_slot),get_burned]),(
10,[touch(bread_slot),get_burned])]
WITH INJURIES: [(1,[insert(bread_slice,bread_slot)]
),(2,[press(on_lever)]),(3,[raise_temp(bread_slot),
touch(bread_slot),get_burned]),(4,[raise_temp(bread
_slot),touch(bread_slot),get_burned]),(5,[raise_tem
p(bread_slot),touch(bread_slot),get_burned]),(6,[ra
ise_temp(bread_slot),touch(bread_slot),get_burned])
,(7,[raise_temp(bread_slice),touch(bread_slot),get_
burned]),(8,[raise_temp(bread_slice),touch(bread_sl
ot),get_burned]),(9,[raise_temp(bread_slice),touch(
bread_slot),get_burned]),(10,[raise_temp(bread_slic
e),touch(bread_slot),get_burned]),(11,[pop_up]),(12
,[remove(bread_slice,bread_slot)])]
Making interpretations...
INTERPRETATIONS: [(1,insert(bread_slice,bread_slot)
),(2,press(on_lever)),(3,raise_temp(bread_slot)),(3
,touch(bread_slot)),(3,get_burned),(3,raise_temp(br
ead_slot)),(3,raise_temp(bread_slot)),(3,raise_temp
(bread_slot)),(3,raise_temp(bread_slice)),(3,raise_
temp(bread_slice)),(3,raise_temp(bread_slice)),(3,r
aise_temp(bread_slice)),(4,pop_up),(5,remove(bread_
slice,bread_slot))]
PATTERNS: [(3,heating_period)]
Making SPL...
Done.
spawn ~/penman/penman
Allegro CL 4.2 [SPARC; R1] (2/3/95 0:50)
Copyright (C) 1985-1993, Franz Inc., Berkeley, CA, USA.  All Rights Reserved.
;; Optimization settings: safety 1, space 1, speed 1, debug 2
;; For a complete description of all compiler switches given the current
;; optimization settings evaluate (EXPLAIN-COMPILER-SETTINGS).
USER(1): :pa penman
#<The PENMAN package>
PENMAN(2): :ld lexicon
; Loading /homes/neat/a/da/thesis/penman/lexicon.
PENMAN(3): :ld comb.loom
; Loading /homes/neat/a/da/thesis/penman/comb.loom.
.+.+.+.+.+-.+.+.+.+.+.+..++..++...++...++..++..++..
++..++..++...++..+..++..+..++..+..++..+..++...++..+
+..++

Warning # 31   --   The SPL macro SPEECHACT is being redefined.
PENMAN(4): :ld toast.spl
; Loading /homes/neat/a/da/thesis/penman/toast.spl.
PENMAN(5): (dolist (x plan)(print (say-spl x)))

"Insert the bread slice into the toaster's bread slot."
"Press the ON lever."
"Do not touch the toaster's bread slot during the heating period."
"The bread slice will pop up."
"Take the bread slice out of the toaster's bread slot."
\end{verbatim}
\end{footnotesize}

\section{Output for the breadmaker domain}
\vspace{4mm}
\begin{footnotesize}
\begin{verbatim}
spawn prolog
Quintus Prolog Release 3.2 (Sun 4, SunOS 5.3)
Copyright (C) 1994, Quintus Corporation.  All rights reserved.
301 East Evelyn Ave, Mountain View, California U.S.A. (415) 254-2800
Licensed to Dept. of Computer Science, University of Toronto

| ?- ['~/prolog/bread'].
% compiling file /homes/neat/a/da/prolog/bread.pl
% bread.pl compiled in module user, 1.410 sec 21,316 bytes

yes
| ?- run.
GOAL STATE: do(remove(bread,baking_pan),do(remove(b
aking_pan,main_body_interior),do(open(lid),do(flash
(complete_light),do(raise_temp(baking_pan),do(raise
_temp(baking_pan),do(steamify(baking_pan),do(raise_
temp(baking_pan),do(raise_temp(baking_pan),do(raise
_temp(steam_vent),do(raise_temp(steam_vent),do(rais
e_temp(steam_vent),do(raise_temp(steam_vent),do(rai
se_temp(main_body),do(raise_temp(main_body),do(stea
mify(main_body),do(raise_temp(main_body),do(raise_t
emp(main_body),do(press(breadmaker_on_button),do(cl
ose(lid),do(insert(baking_pan,main_body_interior),d
o(open(lid),do(pour(yeast,baking_pan_interior),do(p
our(flour,baking_pan_interior),do(pour(water,baking
_pan_interior),do(attach(kneading_blade,baking_pan)
,s0))))))))))))))))))))))))))
Inserting injuries...
INDEX: [1] [2] [3] [4] [5] [6] [7] [8] [9] [10] [11
] [12] [13] [14] [15] [16] [17] [18] [19] [20] [21]
[22] [23] [24] [25] [26]
POINTS: [(10,[touch(main_body),get_burned]),(11,[to
uch(main_body),get_burned]),(12,[touch(main_body),g
et_burned]),(13,[touch(main_body),get_burned]),(14,
[touch(main_body),get_burned]),(15,[touch(steam_ven
t),get_burned]),(16,[touch(steam_vent),get_burned])
,(17,[touch(steam_vent),get_burned]),(18,[touch(ste
am_vent),get_burned]),(19,[touch(steam_vent),get_bu
rned]),(20,[touch(steam_vent),get_burned]),(21,[tou
ch(steam_vent),get_burned]),(22,[touch(steam_vent),
get_burned])]
WITH INJURIES: [(1,[attach(kneading_blade,baking_pa
n)]),(2,[pour(water,baking_pan_interior)]),(3,[pour
(flour,baking_pan_interior)]),(4,[pour(yeast,baking
_pan_interior)]),(5,[open(lid)]),(6,[insert(baking_
pan,main_body_interior)]),(7,[close(lid)]),(8,[pres
s(breadmaker_on_button)]),(9,[raise_temp(main_body)
]),(10,[raise_temp(main_body),touch(main_body),get_
burned]),(11,[steamify(main_body),touch(main_body),
get_burned]),(12,[raise_temp(main_body),touch(main_
body),get_burned]),(13,[raise_temp(main_body),touch
(main_body),get_burned]),(14,[raise_temp(steam_vent
),touch(main_body),get_burned]),(15,[raise_temp(ste
am_vent),touch(steam_vent),get_burned]),(16,[raise_
temp(steam_vent),touch(steam_vent),get_burned]),(17
,[raise_temp(steam_vent),touch(steam_vent),get_burn
ed]),(18,[raise_temp(baking_pan),touch(steam_vent),
get_burned]),(19,[raise_temp(baking_pan),touch(stea
m_vent),get_burned]),(20,[steamify(baking_pan),touc
h(steam_vent),get_burned]),(21,[raise_temp(baking_p
an),touch(steam_vent),get_burned]),(22,[raise_temp(
baking_pan),touch(steam_vent),get_burned]),(23,[fla
sh(complete_light)]),(24,[open(lid)]),(25,[remove(b
aking_pan,main_body_interior)]),(26,[remove(bread,b
aking_pan)])]
Making interpretations...
INTERPRETATIONS: [(1,attach(kneading_blade,baking_p
an)),(2,pour(water,baking_pan_interior)),(3,pour(fl
our,baking_pan_interior)),(4,pour(yeast,baking_pan_
interior)),(5,open(lid)),(6,insert(baking_pan,main_
body_interior)),(7,close(lid)),(8,press(breadmaker_
on_button)),(9,raise_temp(main_body)),(9,raise_temp
(main_body)),(9,touch(main_body)),(9,get_burned),(9
,steamify(main_body)),(9,raise_temp(main_body)),(9,
raise_temp(main_body)),(9,raise_temp(steam_vent)),(
9,raise_temp(steam_vent)),(9,touch(steam_vent)),(9,
get_burned),(9,raise_temp(steam_vent)),(9,raise_tem
p(steam_vent)),(9,raise_temp(baking_pan)),(9,raise_
temp(baking_pan)),(9,steamify(baking_pan)),(9,raise
_temp(baking_pan)),(9,raise_temp(baking_pan)),(10,f
lash(complete_light)),(11,open(lid)),(12,remove(bak
ing_pan,main_body_interior)),(13,remove(bread,bakin
g_pan))]
PATTERNS: [(9,heating_period)]
Making SPL...
Done.
spawn ~/penman/penman
Allegro CL 4.2 [SPARC; R1] (2/3/95 0:50)
Copyright (C) 1985-1993, Franz Inc., Berkeley, CA, USA.  All Rights Reserved.
;; Optimization settings: safety 1, space 1, speed 1, debug 2
;; For a complete description of all compiler switches given the current
;; optimization settings evaluate (EXPLAIN-COMPILER-SETTINGS).
USER(1): :pa penman
#<The PENMAN package>
PENMAN(2): :ld lexicon
; Loading /homes/neat/a/da/thesis/penman/lexicon.
PENMAN(3): :ld comb.loom
; Loading /homes/neat/a/da/thesis/penman/comb.loom.
.+.+.+.+.+-.+.+.+.+.+.+..++..++...++...++..++..++..
++..++..++...++..+..++..+..++..+..++..+..++...++..+
+..++

Warning # 31   --   The SPL macro SPEECHACT is being redefined.
PENMAN(4): :ld bread.spl
; Loading /homes/neat/a/da/thesis/penman/bread.spl.
PENMAN(5): (dolist (x plan)(print (say-spl x)))

"Attach the kneading blade to the baking pan."
"Pour the water into the baking pan."
"Pour the flour into the baking pan."
"Pour the yeast into the baking pan."
"Open the lid."
"Insert the baking pan into the main body."
"Close the lid."
"Press the ON button."
"Do not touch the main body during the heating period."
"Do not touch the steam vent during the heating period."
"The ``complete'' light will flash."
"Open the lid."
"Take the baking pan out of the main body."
"Take the bread from the baking pan."
\end{verbatim}
\end{footnotesize}

\section{Output for the breadmaker/toaster combination domain}
\vspace{4mm}
\begin{footnotesize}
\begin{verbatim}
spawn prolog
Quintus Prolog Release 3.2 (Sun 4, SunOS 5.3)
Copyright (C) 1994, Quintus Corporation.  All rights reserved.
301 East Evelyn Ave, Mountain View, California U.S.A. (415) 254-2800
Licensed to Dept. of Computer Science, University of Toronto

| ?- ['~/prolog/comb'].
% compiling file /homes/neat/a/da/prolog/comb.pl
% comb.pl compiled in module user, 1.630 sec 23,528 bytes

yes
| ?- run.
GOAL STATE: do(remove(bread_slice,bread_slot),do(po
p_up,do(raise_temp(bread_slice),do(raise_temp(bread
_slice),do(raise_temp(bread_slice),do(raise_temp(br
ead_slice),do(raise_temp(bread_slot),do(raise_temp(
bread_slot),do(raise_temp(bread_slot),do(raise_temp
(bread_slot),do(press(on_lever),do(insert(bread_sli
ce,bread_slot),do(slice(bread_slice,bread),do(remov
e(bread,baking_pan_interior),do(remove(baking_pan,m
ain_body_interior),do(open(lid),do(flash(complete_l
ight),do(raise_temp(steam_vent),do(raise_temp(steam
_vent),do(raise_temp(steam_vent),do(raise_temp(stea
m_vent),do(raise_temp(main_body),do(raise_temp(main
_body),do(steamify(main_body),do(raise_temp(main_bo
dy),do(raise_temp(main_body),do(press(breadmaker_on
_button),do(close(lid),do(insert(baking_pan,main_bo
dy_interior),do(open(lid),do(pour(yeast,baking_pan_
interior),do(pour(flour,baking_pan_interior),do(pou
r(water,baking_pan_interior),do(attach(kneading_bla
de,baking_pan),s0))))))))))))))))))))))))))))))))))
Inserting injuries...
INDEX: [1] [2] [3] [4] [5] [6] [7] [8] [9] [10] [11
] [12] [13] [14] [15] [16] [17] [18] [19] [20] [21]
[22] [23] [24] [25] [26] [27] [28] [29] [30] [31] [
32] [33] [34]
POINTS: [(10,[touch(main_body),get_burned]),(11,[to
uch(main_body),get_burned]),(12,[touch(main_body),g
et_burned]),(13,[touch(main_body),get_burned]),(14,
[touch(main_body),get_burned]),(15,[touch(steam_ven
t),get_burned]),(16,[touch(steam_vent),get_burned])
,(17,[touch(steam_vent),get_burned]),(26,[touch(bre
ad_slot),get_burned]),(27,[touch(bread_slot),get_bu
rned]),(28,[touch(bread_slot),get_burned]),(29,[tou
ch(bread_slot),get_burned]),(30,[touch(bread_slot),
get_burned]),(31,[touch(bread_slot),get_burned]),(3
2,[touch(bread_slot),get_burned])]
WITH INJURIES: [(1,[attach(kneading_blade,baking_pa
n)]),(2,[pour(water,baking_pan_interior)]),(3,[pour
(flour,baking_pan_interior)]),(4,[pour(yeast,baking
_pan_interior)]),(5,[open(lid)]),(6,[insert(baking_
pan,main_body_interior)]),(7,[close(lid)]),(8,[pres
s(breadmaker_on_button)]),(9,[raise_temp(main_body)
]),(10,[raise_temp(main_body),touch(main_body),get_
burned]),(11,[steamify(main_body),touch(main_body),
get_burned]),(12,[raise_temp(main_body),touch(main_
body),get_burned]),(13,[raise_temp(main_body),touch
(main_body),get_burned]),(14,[raise_temp(steam_vent
),touch(main_body),get_burned]),(15,[raise_temp(ste
am_vent),touch(steam_vent),get_burned]),(16,[raise_
temp(steam_vent),touch(steam_vent),get_burned]),(17
,[raise_temp(steam_vent),touch(steam_vent),get_burn
ed]),(18,[flash(complete_light)]),(19,[open(lid)]),
(20,[remove(baking_pan,main_body_interior)]),(21,[r
emove(bread,baking_pan_interior)]),(22,[slice(bread
_slice,bread)]),(23,[insert(bread_slice,bread_slot)
]),(24,[press(on_lever)]),(25,[raise_temp(bread_slo
t)]),(26,[raise_temp(bread_slot),touch(bread_slot),
get_burned]),(27,[raise_temp(bread_slot),touch(brea
d_slot),get_burned]),(28,[raise_temp(bread_slot),to
uch(bread_slot),get_burned]),(29,[raise_temp(bread_
slice),touch(bread_slot),get_burned]),(30,[raise_te
mp(bread_slice),touch(bread_slot),get_burned]),(31,
[raise_temp(bread_slice),touch(bread_slot),get_burn
ed]),(32,[raise_temp(bread_slice),touch(bread_slot)
,get_burned]),(33,[pop_up]),(34,[remove(bread_slice
,bread_slot)])]
Making interpretations...
INTERPRETATIONS: [(1,attach(kneading_blade,baking_p
an)),(2,pour(water,baking_pan_interior)),(3,pour(fl
our,baking_pan_interior)),(4,pour(yeast,baking_pan_
interior)),(5,open(lid)),(6,insert(baking_pan,main_
body_interior)),(7,close(lid)),(8,press(breadmaker_
on_button)),(9,raise_temp(main_body)),(9,raise_temp
(main_body)),(9,touch(main_body)),(9,get_burned),(9
,steamify(main_body)),(9,raise_temp(main_body)),(9,
raise_temp(main_body)),(9,raise_temp(steam_vent)),(
9,raise_temp(steam_vent)),(9,touch(steam_vent)),(9,
get_burned),(9,raise_temp(steam_vent)),(9,raise_tem
p(steam_vent)),(10,flash(complete_light)),(11,open(
lid)),(12,remove(baking_pan,main_body_interior)),(1
3,remove(bread,baking_pan_interior)),(14,slice(brea
d_slice,bread)),(15,insert(bread_slice,bread_slot))
,(16,press(on_lever)),(17,raise_temp(bread_slot)),(
17,raise_temp(bread_slot)),(17,touch(bread_slot)),(
17,get_burned),(17,raise_temp(bread_slot)),(17,rais
e_temp(bread_slot)),(17,raise_temp(bread_slice)),(1
7,raise_temp(bread_slice)),(17,raise_temp(bread_sli
ce)),(17,raise_temp(bread_slice)),(18,pop_up),(19,r
emove(bread_slice,bread_slot))]
PATTERNS: [(9,heating_period),(17,heating_period)]
Making SPL...
Done.
spawn ~/penman/penman
Allegro CL 4.2 [SPARC; R1] (2/3/95 0:50)
Copyright (C) 1985-1993, Franz Inc., Berkeley, CA, USA.  All Rights Reserved.
;; Optimization settings: safety 1, space 1, speed 1, debug 2
;; For a complete description of all compiler switches given the current
;; optimization settings evaluate (EXPLAIN-COMPILER-SETTINGS).
USER(1): :pa penman
#<The PENMAN package>
PENMAN(2): :ld lexicon
; Loading /homes/neat/a/da/thesis/penman/lexicon.
PENMAN(3): :ld comb.loom
; Loading /homes/neat/a/da/thesis/penman/comb.loom.
.+.+.+.+.+-.+.+.+.+.+.+..++..++...++...++..++..++..
++..++..++...++..+..++..+..++..+..++..+..++...++..+
+..++

Warning # 31   --   The SPL macro SPEECHACT is being redefined.
PENMAN(4): :ld comb.spl
; Loading /homes/neat/a/da/thesis/penman/comb.spl.
PENMAN(5): (dolist (x plan)(print (say-spl x)))

"Attach the kneading blade to the baking pan."
"Pour the water into the baking pan."
"Pour the flour into the baking pan."
"Pour the yeast into the baking pan."
"Open the lid."
"Insert the baking pan into the main body."
"Close the lid."
"Press the ON button."
"Do not touch the main body during the heating period."
"Do not touch the steam vent during the heating period."
"The ``complete'' light will flash."
"Open the lid."
"Take the baking pan out of the main body."
"Take the bread out of the baking pan."
"Cut the bread slice from the bread."
"Insert the bread slice into the toaster's bread slot."
"Press the ON lever."
"Do not touch the toaster's bread slot during the heating period."
"The bread slice will pop up."
"Take the bread slice out of the toaster's bread slot."
\end{verbatim}
\end{footnotesize}

\chapter{The SPL files}

This appendix contains the SPL corresponding to the instructions
generated for the toaster and breadmaker domains.  The SPL for the
toaster/breadmaker combination is very similar to the concatenation of
the SPL for the toaster and breadmaker domains; it has been omitted for
this reason.

\section{SPL for the toaster instructions}
\vspace{4mm}
\begin{footnotesize}
\begin{verbatim}
(setq plan '((ID1 / insert
        :actor (hearer / person)
        :actee (ID2 / bread_slice
                :determiner the)
        :destination (ID3 / bread_slot
                :determiner the)
        :tense present
        :speechact imperative)
(ID4 / press
        :actor (hearer / person)
        :actee (ID5 / on_lever
                :determiner the)
        :tense present
        :speechact imperative)
(ID6 / touch
        :actor (hearer / person)
        :actee (ID7 / bread_slot
                :determiner the)
        :exhaustive-duration (ID8 / heating_period
                :determiner the)
        :tense present
        :speechact neg-imperative)
(ID9 / pop_up
        :actor (ID10 / bread_slice
                :determiner the)
        :tense future
        :speechact assertion)
(ID11 / remove
        :actor (hearer / person)
        :actee (ID12 / bread_slice
                :determiner the)
        :source (ID13 / bread_slot
                :determiner the)
        :tense present
        :speechact imperative)))
\end{verbatim}
\end{footnotesize}

\section{SPL for the breadmaker instructions}
\vspace{4mm}
\begin{footnotesize}
\begin{verbatim}
(setq plan '((ID1 / attach
        :actor (hearer / person)
        :actee (ID2 / kneading_blade
                :determiner the)
        :destination (ID3 / baking_pan
                :determiner the)
        :tense present
        :speechact imperative)
(ID4 / pour
        :actor (hearer / person)
        :actee (ID5 / water
                :determiner the)
        :destination (ID6 / baking_pan_interior
                :determiner the)
        :tense present
        :speechact imperative)
(ID7 / pour
        :actor (hearer / person)
        :actee (ID8 / flour
                :determiner the)
        :destination (ID9 / baking_pan_interior
                :determiner the)
        :tense present
        :speechact imperative)
(ID10 / pour
        :actor (hearer / person)
        :actee (ID11 / yeast
                :determiner the)
        :destination (ID12 / baking_pan_interior
                :determiner the)
        :tense present
        :speechact imperative)
(ID13 / open
        :actor (hearer / person)
        :actee (ID14 / lid
                :determiner the)
        :tense present
        :speechact imperative)
(ID15 / insert
        :actor (hearer / person)
        :actee (ID16 / baking_pan
                :determiner the)
        :destination (ID17 / main_body_interior
                :determiner the)
        :tense present
        :speechact imperative)
(ID18 / close
        :actor (hearer / person)
        :actee (ID19 / lid
                :determiner the)
        :tense present
        :speechact imperative)
(ID20 / press
        :actor (hearer / person)
        :actee (ID21 / breadmaker_on_button
                :determiner the)
        :tense present
        :speechact imperative)
(ID22 / touch
        :actor (hearer / person)
        :actee (ID23 / main_body
                :determiner the)
        :exhaustive-duration (ID24 / heating_period
                :determiner the)
        :tense present
        :speechact neg-imperative)
(ID25 / touch
        :actor (hearer / person)
        :actee (ID26 / steam_vent
                :determiner the)
        :exhaustive-duration (ID27 / heating_period
                :determiner the)
        :tense present
        :speechact neg-imperative)
(ID28 / flash
        :actor (ID29 / complete_light
                :determiner the)
        :tense future
        :speechact assertion)
(ID30 / open
        :actor (hearer / person)
        :actee (ID31 / lid
                :determiner the)
        :tense present
        :speechact imperative)
(ID32 / remove
        :actor (hearer / person)
        :actee (ID33 / baking_pan
                :determiner the)
        :source (ID34 / main_body_interior
                :determiner the)
        :tense present
        :speechact imperative)
(ID35 / remove
        :actor (hearer / person)
        :actee (ID36 / bread
                :determiner the)
        :source (ID37 / baking_pan
                :determiner the)
        :tense present
        :speechact imperative)))
\end{verbatim}
\end{footnotesize}

\bibliographystyle{authdate}

\end{document}